\pdfoutput=1
\documentclass[journal,twoside,web]{ieeecolor}

\usepackage{amsmath}
\usepackage{amsfonts,amssymb,bm,fixmath}
\usepackage{algorithm2e}
\usepackage{hyperref}
\usepackage{url}
\usepackage{graphicx}
\usepackage{float}
\usepackage{caption}
\usepackage{tikz}
\usetikzlibrary{arrows}
\usetikzlibrary{fit}
\usepackage{xcolor}
\usepackage{multirow}
\usepackage{subcaption}
\usepackage{wrapfig}
\usepackage{mathtools}
\usepackage{ragged2e}
\usepackage[export]{adjustbox}
\usepackage{xr-hyper}
\usepackage{booktabs}
\usepackage{array}
\usepackage[nice]{nicefrac}
\usepackage{xspace}
\usepackage{xcolor}
\usepackage[many]{tcolorbox}
\usepackage{cleveref}

\newtheorem{lemma}{Lemma}[section]
\newtheorem{proposition}{Proposition}[section]

\usepackage{dblfloatfix}    %
\usepackage[bottom]{footmisc} %

\usepackage{cite}
\usepackage{algorithmic}
\usepackage{graphicx}
\usepackage{textcomp}
\usepackage[outdir=./]{epstopdf}

\def\eqref#1{equation~\ref{#1}}
\def\Eqref#1{Equation~\ref{#1}}

\def\1{\bm{1}}

\def\rd{{\textnormal{d}}}

\def\vzero{{\bm{0}}}

\def\vmu{{\bm{\mu}}}

\def\vc{{\bm{c}}}

\def\vf{{\bm{f}}}
\def\vg{{\bm{g}}}
\def\vh{{\bm{h}}}

\def\vk{{\bm{k}}}

\def\vq{{\bm{q}}}

\def\vu{{\bm{u}}}
\def\vv{{\bm{v}}}

\def\vx{{\bm{x}}}
\def\vy{{\bm{y}}}
\def\vz{{\bm{z}}}

\def\mA{{\bm{A}}}
\def\mB{{\bm{B}}}
\def\mC{{\bm{C}}}

\def\mF{{\bm{F}}}

\def\mH{{\bm{H}}}
\def\mI{{\bm{I}}}
\def\mJ{{\bm{J}}}
\def\mK{{\bm{K}}}

\def\mP{{\bm{P}}}
\def\mQ{{\bm{Q}}}

\def\mS{{\bm{S}}}

\def\mU{{\bm{U}}}
\def\mV{{\bm{V}}}

\def\mX{{\bm{X}}}

\def\mZ{{\bm{Z}}}

\def\mLambda{{\bm{\Lambda}}}
\def\mSigma{{\bm{\Sigma}}}

\DeclareMathAlphabet{\mathsfit}{\encodingdefault}{\sfdefault}{m}{sl}
\SetMathAlphabet{\mathsfit}{bold}{\encodingdefault}{\sfdefault}{bx}{n}

\def\gN{{\mathcal{N}}}
\def\gO{{\mathcal{O}}}

\def\sR{{\mathbb{R}}}

\newcommand{\E}{\mathbb{E}}

\DeclareMathOperator*{\argmin}{arg\,min}

\usepackage{amssymb}
\usepackage{fixmath}

\newcommand{\mum}[3]{\mathbold{\mu}_{{#1}_{#3}^{#2}}}

\newcommand{\sigm}[3]{\mathbold{\Sigma}_{{#1}_{#3}^{#2}}}

\def\mfSigma{\overrightarrow{{\bm{\Sigma}}}}
\def\mbSigma{\overleftarrow{{\bm{\Sigma}}}}
\def\mfLambda{\overrightarrow{{\bm{\Lambda}}}}
\def\mbLambda{\overleftarrow{{\bm{\Lambda}}}}

\newcommand{\mat}[1]{\mathbold{#1}}
\newcommand{\vtau}{\mat{\tau}}
\newcommand{\vpi}{\mat{\pi}}
\newcommand{\vnu}{\mat{\nu}}
\newcommand{\vzeta}{\mat{\zeta}}
\newcommand{\veta}{\mat{\eta}}
\newcommand{\vxi}{\mat{\xi}}

\newcommand{\mTheta}{\mat{\Theta}}

\newcommand{\sigEta}[1]{\mat{\Sigma}_{\veta_{#1}}}
\newcommand{\sigXi}{\mat{\Sigma}_{\vxi}}

\newcommand{\tran}{^\top}
\newcommand{\inv}{^{-1}}
\newcommand{\mahalanobis}[2]{\lVert #1\rVert_{#2}^2}

\definecolor{color0}{rgb}{0,0.75,0.75}
\definecolor{color1}{rgb}{0,0.75,0.75}

\definecolor{yellow}{rgb}{0.75,0.75,0}
\definecolor{magenta}{rgb}{0.75,0,0.75}
\definecolor{cyan}{rgb}{0,0.75,0.75}
\definecolor{green}{rgb}{0.0,0.50,0.25}
\newcommand{\itwoc}{%
	\textsc{i2c}\xspace
}

\usepackage{pgfplots}
\pgfplotsset{compat=newest}
\usepgfplotslibrary{groupplots}
\usepgfplotslibrary{dateplot}

\DeclareMathOperator{\tr}{tr}

\newtcbtheorem[auto counter]{definition}{definition}{
	lower separated=false,
	colback=white,
	fonttitle=\sc,
	colbacktitle=white,
	coltitle=subsectioncolor,
	colframe=subsectioncolor,
	enhanced,
	boxed title style={colback=white,opacityback=1, opacityframe=0}, %
	boxed title style={colframe=subsectioncolor},
	attach boxed title to top left={xshift=0.5cm,yshift=-3mm},
}{def}

\newtcolorbox{figeq}[2][]{
	rounded corners=all,
	enhanced,
	colback=white!90!gray,
	coltitle=subsectioncolor,
	colbacktitle=white!90!gray,
	fonttitle=\sc,
	frame hidden,
	attach boxed title to top left={yshift=-3.5mm},
	boxed title style={opacityback=0., opacityframe=0.},
	title=#2,#1
}

\usepackage{generic}

\def\BibTeX{{\rm B\kern-.05em{\sc i\kern-.025em b}\kern-.08em
    T\kern-.1667em\lower.7ex\hbox{E}\kern-.125emX}}
\begin{document}

\title{Efficient Stochastic Optimal Control through\\Approximate Bayesian Input Inference}
\author{
	Joe Watson,
	Hany Abdulsamad,
	Rolf Findeisen, \IEEEmembership{Member, IEEE}, and
	Jan Peters, \IEEEmembership{Fellow, IEEE}
	\thanks{
	This project has received funding from the European Union’s Horizon 2020 program under grant agreement No \#640554 (SKILLS4ROBOTS).
	J. Watson and J. Peters are with the Intelligent Autonomous Systems group, Technical University of Darmstadt.
	H. Abdulsamad is with the Finnish Centre of AI and the Electrical Engineering and Automation Department, Aalto University.
	R. Findeisen is with the Control and Cyber-Physical Systems Laboratory, Technical University of Darmstadt. Contact: \texttt{\{joe, hany,jan\}@robot-learning.de, rolf.findeisen@tu-darmstadt.de.}}
}

\maketitle

\begin{abstract}
	Optimal control under uncertainty is a prevailing challenge for many reasons. One of the critical difficulties lies in producing tractable solutions for the underlying stochastic optimization problem.
	We show how advanced approximate inference techniques can be used to handle the statistical approximations principled and practically by framing the control problem as a problem of input estimation.
	Analyzing the Gaussian setting, we present an inference-based solver that is effective in stochastic and deterministic settings and was found to be superior to popular baselines on nonlinear simulated tasks.
	We draw connections that relate this inference formulation to previous approaches for stochastic optimal control and outline several advantages that this inference view brings due to its statistical nature.
\end{abstract}


\vspace{-1em}
\section{Introduction}
Control-as-inference \cite{Dayan97,Attias03planningby,06-toussaint-ICML,kappen2013} refers to the formulation of optimal control as inference of an equivalent probabilistic graphical model. 
The motivation for this perspective is threefold:
Firstly, it allows one to derive intriguing mathematical dualities between the field of control and probabilistic inference, which has captivated researchers for decades \cite{kalman1960new,todorov2008general,TheodorouT12}.
Secondly, in making the effort to reframe the optimal control problem as one of Bayesian inference, we gain access to a sophisticated suite of tools and insights developed by the statistics community that provides value in both theory and practice \cite{hennig2015probabilistic}.
In particular, as stochasticity presents a challenge in designing tractable algorithms for control under uncertainty \cite{7740982}, approximate inference techniques \cite{barberBRML2011} provide means of obtaining distributions through principled approximations.   
Finally, an abundance of data, combined with the pervasive demand for sophisticated control design, has driven the study of \emph{learning} for control, where experience is leveraged within the control optimization process \cite{matni2019self, hewing2020learning, maiworm2021online}.
In this setting, the motivation of inference-based optimal control is clear, as an inference-based formulation is complementary to a learned probabilistic dynamics model.
Using the language of inference, optimal control and data can be integrated directly to synthesize effective control learning algorithms~\cite{deisenroth2013survey}.

Taking the inference perspective, this article concerns the application of Bayesian smoothing methods for optimal control, focusing on the Gaussian setting \cite{toussaint2009robot, i2corl, i2cacc}.
Specifically, this work builds on the view of optimal control as the problem of input estimation \cite{hoffmann2017linear}, developing the seminal work reformulating approximate optimal control as approximate message passing on a probabilistic graphical model \cite{toussaint2009robot, rawlik2013probabilistic}.
Rather than using the regularized Riccati equation updates \cite{toussaint2009robot}, we leverage the inference duality and perform optimization as a Bayesian smoothing problem in the state-action space.
This view provides an alternative to linearization-based nonlinear optimal control and yields accurate, regularized estimates of the value functions using approximate inference.
Moreover, inspired by inference of state-space models, we propose an expectation maximization scheme that enables the optimization of priors and hyperparameters, which is beneficial when applying these methods to complex control tasks which often require iterative optimization in a non-convex setting.
In contrast, previous approaches are limited to fixed priors and hyperparameters, estimating the posterior state distribution using iterative relinearization of the dynamics
~\cite{toussaint2009robot, rawlik2013probabilistic}.

\subsection{Contribution}
Prior work proposed input inference for optimal control using open-loop optimization with linearization \cite{i2corl} and closed-loop optimization with expert controllers and covariance control \cite{i2cacc}. 
This work provides a comprehensive review of the approach, with a novel analysis of the Gaussian control-inference duality through the posterior covariances and state-action value functions.
Moreover, we compare the performance of different approximate inference techniques, and outline applications of our approach for high-dimensional and partially-observed model predictive control.
We also discuss connections to risk-sensitive, maximum entropy and dual control; as well as optimization and exploration perspectives.

\subsection{Related Work}
\label{sec:related}
Control-as-inference has been considered since the conception of modern optimal control, due to the concurrent development of linear quadratic regulator (LQR), Kalman filtering and linear quadratic regulator (LQG) \cite{kalman1960new, bayesian_control}.\\
The revival in control-as-inference stems from simultaneous works that tackle the setting from different perspectives.
Path integral control \cite{kappen2013, kappen2016adaptive, williams2017model} exploits the duality between integration over trajectories and Monte Carlo expectations to derive a sample-based optimal control scheme that performs adaptive importance sampling via stochastic perturbations.
Linearly-solvable Markov decision processes \cite{todorov2007linearly} present a closely related approach, but in a discrete state space, where the disturbance assumption allows the problem to be solved as a linear program.
Trans-dimensional Markov chain Monte Carlo \cite{NIPS2007_3a15c7d0} provides a principled means of performing sample-based finite-horizon optimal control that can jointly optimize the task horizon. 
Message passing methods \cite{Attias03planningby, 06-toussaint-ICML,toussaint2009robot,rawlik2013stochastic} allow the direct duality to be established between inference and dynamic programming-based control in discrete and continuous state spaces.
In the Gaussian setting, there is a strong correspondence to the LQG problem \cite{06-toussaint-ICML,toussaint2009robot}, and so approximate Gaussian inference similarly derives approximate inference control (AICO), an open-loop regularized Riccatti equations nonlinear solver that resembles differential dynamic programming \cite{jacobson1970differential}, specifically the Gauss-Newton approximation, i.e. iterative LQR (iLQR) \cite{li2004iterative,todorov2005generalized}.
For nonlinear smoothing, AICO uses iterated inference, relinearizing about the posterior state mode each iteration.
Posterior policy iteration (PPI) \cite{rawlik2013probabilistic} extends AICO for feedback control, framing a linear Gaussian stochastic controller as a conditional distribution such that the posterior controls can be computed using the same iterated inference.

The inference perspective for control is strongly motivated by the reinforcement learning setting, which uses both optimal control and statistical methods \cite{Dayan97,neumann_icml2011,levine2018reinforcement,o2018variational}.
Furthermore, trajectory optimization has been regularized using information-geometric constraints, such as the Kullback-Leibler (KL) divergence \cite{Levine_gps, lioutikov_icra_2014,Abdulsamad_ICAPS_2017}, which can be interpreted as variational inference \cite{levine2013variational}.
Inference methods have also been adopted for motion planning \cite{7989082}, applying optimized factor graph solvers to planning tasks~\cite{Mukadam_mp}. 

Conversely, many techniques popularized by probabilistic inference have been applied independently to enhance optimal control.
Extended LQR incorporates a filtering-like forward optimization into iLQR,\cite{van2016extended, sun_elqr}.  
Quadrature methods have been adopted for trajectory optimization for greater accuracy \cite{4927530, NIPS2007_c6bff625, 7798817, 9387078}.
Sampled differential dynamic programming uses Monte Carlo rollouts to accurately estimate value functions of non-smooth dynamics, by using relating the sample covariance to the inverse Hessian of the log-likelihood objective \cite{saDDP1,saDDP2}.
Inference-based methods have also been used for model predictive control.
Sequential Monte Carlo was used a sample-based solver for non-convex, non-Gaussian MPC \cite{kantas2009sequential}.
Stein variational gradient descent has also been used to combine sample- and gradient-based computation for MPC to plan over multi-modal trajectories~\cite{lambert2020stein}.

\subsection{Notation}
The matrix $\mX_1^T$ denotes a sequence of vectors $\{\vx_1,\dots,\vx_T\}$.
For vectors $\vx_1$, $\vx_2$ and pos. def. matrix $\mS{\,\succ\,}0$ we define
$\mahalanobis{\vx_1{-}\vx_2}{\mS}{=}\textstyle\frac{1}{2}(\vx_1{-}\vx_2)\tran\mS(\vx_1{-}\vx_2)$.
The probability of an event $X$ is denoted by $\text{Pr}(X)$.
The expression $\vx{\,\sim\,}p(\vx)$  means that a random variable $\vx{\,\in\,}\sR^n$ is distributed according to the density $p(\vx)$.
The log-likelihood under this distribution is expressed as $\mathcal{L}(\vx){\,=\,}\log p(\vx)$.
A multivariate Normal distribution with mean $\vmu\in\sR^n$ and covariance $\mSigma{\,\in\,}S^n_+$ is $\gN(\vmu,\mSigma)$, where $+$ denotes positive (semi-) definiteness.
Normal and Gaussian are used interchangeably to refer to this distribution, and it is also presented in its \emph{canonical} form $\gN[\vnu, \mLambda]$, where $\mLambda=\mSigma\inv$ and $\vnu=\mLambda\vmu$.
$\E[\vx]$ and $\mathbb{V}[\vx]$ denote the mean and covariance of $\vx$ respectively.

\section{Stochastic Optimal Control}
\label{sec:background}
Since the offspring of control, optimally controlling systems subject to stochastic disturbances have been examined, see e.g. \cite{charnes1963deterministic, aastrom2012introduction, bryson2018applied, mesbah2014stochastic, paulson2014fast}. 
In this section, we review relevant topics of stochastic optimal control (SOC) \cite{stengel1986stochastic}.
These topics are outlined in order to identify how inference can be used to perform optimal control.
Specifically, we connect risk-sensitive linear quadratic dynamic programming to linear Gaussian Bayesian smoothing.

\subsection{Finite-Horizon, Discrete-Time Optimal Control}
\label{sec:soc}
We consider control of a stochastic, discrete-time, fully-observed, nonlinear, time-varying dynamical system, $\vf_t$, with state $\vx \in \mathbb{R}^{d_x}$ and input $\vu \in \mathbb{R}^{d_u}$.
For such a system, we desire the optimal controls over a horizon of $T$ time steps that minimizes the time-varying cost functions $C_t:\sR^{d_x}\times\sR^{d_u}\!\rightarrow\!\sR$ in expectation over the dynamics from an initial state $\vx_0$,
\begin{align}
	&\textstyle\min_{\mU_1^{T-1}}& &\mathbb{E}[C_T(\vx_T) + \textstyle\sum_{t=1}^{T-1}\;C_t(\vx_t, \vu_t)]\hspace{1cm}\notag\\
	&\text{s.t.}& &\vx_{t+1} = \vf_t(\vx_t, \vu_t)+\veta_{t},\;\veta_{t}\sim\gN(\vzero,\sigEta{t}).\label{eq:fhoc}
\end{align}
Various approaches to exactly or approximately solve this problem exist \cite{bertsekas1996stochastic,stengel1986stochastic}. 
The dynamic programming solution \cite{bertsekas1996stochastic} introduces the state and state-action value functions $V$ and $Q$, where
\begin{align}
	V_t(\vx) &=
	\min_{\vu_t}Q_t(\vx_t, \vu_t),\;V_T(\vx_T) = C_T(\vx_T), \text{ and} \label{eq:bellman_eq}\\
	Q_t(\vx_t, \vu_t) &= \mathbb{E}[C_t(\vx_t, \vu_t) + V_{t+1}(\vf_t(\vx_t, \vu_t))].
	\label{eq:q_definition}
\end{align}
We briefly summarize the linear quadratic Gaussian (LQG \cite{bryson2018applied})
, which we will use to identify the inference relations in the linear Gaussian setting.
For compactness, we introduce the extended state-action vector $\vtau$, where $\vtau\!=\![\vx\;\vu]\tran\!\in\!\sR^{d_\tau}$.

In the case of LQR, the cost function is quadratic,
\begin{align}
	C_t(\vx, \vu) &= 
	c_t \!+\!
	\vc_t\tran	\begin{bmatrix}
		\vx\\
		\vu\\
	\end{bmatrix}
	\!+\!
	\begin{bmatrix}
		\vx\\
		\vu\\
	\end{bmatrix}\tran\!
	\begin{bmatrix}
	\mC_{\vx\vx_t}       & \mC_{\vx\vu_t}\\
	\mC_{\vx\vu_t}\tran  & \mC_{\vu\vu_t}\\
	\end{bmatrix}
	\begin{bmatrix}
	\vx\\
	\vu\\
	\end{bmatrix},\notag
	\\
	 &= \vtau\tran\mC_t \vtau + \vc_t\tran\vtau + c_t,\text{ where $\mC_t\succ0\;\forall\;t$.}\notag
\intertext{The state-action value function is also quadratic,}
	Q_t(\vx_t, \vu_t) &= q_t \!+\! \vq_t\tran\!
	\begin{bmatrix}
		\vx_t\\
		\vu_t\\
	\end{bmatrix} {+} 
		\begin{bmatrix}
		\vx_t\\
		\vu_t\\
	\end{bmatrix}\tran\!
	\begin{bmatrix}
		\mQ_{\vx\vx_t} & \hspace{-0.7em}\mQ_{\vx\vu_t} \\
		\mQ_{\vu\vx_t} & \hspace{-0.7em}\mQ_{\vu\vu_t}\\
	\end{bmatrix}\!
	\begin{bmatrix}
		\vx_t\\
		\vu_t\\
	\end{bmatrix}\!\notag
	\\
					  &= q_t + \vq_t\tran\vtau_t + \vtau_t\tran\mQ_{t}\vtau_t,\notag
\intertext{as is the value function, $V_t(\vx_t) = v_t \!+\! \vv_t\tran\vx_t + \vx_t\tran \mV_t \vx_t$. \endgraf
\noindent The dynamics are affine Gaussian in the state,}
	\vx_{t+1} &=\!
	\mF_t\!
	\begin{bmatrix}
	\vx_t\\
	\vu_t\\
	\end{bmatrix}\!
	+ \bar{\vf}_t + \veta_t,\!\text{ where }\!\mF_t\!=\![\mF_{\vx_t}\;\mF_{\vu_t}].\!\!\label{eq:quadratic_v}
\end{align}
The $Q$ function update, (\ref{eq:q_definition}), becomes
\begin{align}
	\vq_t &= \vc_t + \vv_{t+1}\mF_{t+1},\;\;
	\mQ_{t} = \mC_t + \mF_{t+1}\tran \mV_{t+1}\mF_{t+1}. \label{eq:Q_update}
\intertext{Using $Q$, the optimal control input is $\vu_t^*{=}\textstyle\argmin_{\vu_t}\!Q_t(\vx_t, \vu_t)$,}
	\vu_t^* &= -\mQ_{\vu\vu_t}\inv(\vq_{\vu_t} + \mQ_{\vu\vx_t}\vx_t)=\mK_t\vx_t + \vk_t = \vpi_t(\vx_t),\label{eq:optimal_control}
\end{align}
yielding a time-varying linear control law. Substituting equation (\ref{eq:optimal_control}) into (\ref{eq:bellman_eq}), the value function updates are
\begin{align}
	\hspace{-1em}\vv_t\!&=\!\vq_{\vx_t}\!-\!\vq_{\vu_t}\tran\mQ_{\vu\vu_t}\inv\mQ_{\vu\vx_t},\,
	\mV_t\!=\!\mQ_{\vx\vx_t}\!-\!\mQ_{\vx\vu_t}\mQ_{\vu\vu_t}\inv\mQ_{\vu\vx_t}.\!\! \label{eq:V_update}
\end{align}
Similar holds for the so-called differential dynamic programming solution, where the cost and dynamics are updated with a second-order Taylor approximation each iteration.
The assumption of time-varying affine dynamics correspond to the so-called Gauss-Newton approximation \cite{li2004iterative,todorov2005generalized}.

\subsection{Risk-Sensitive Control}
\label{sec:risk}
Introduced by Jacobson, risk-sensitive linear exponential quadratic Gaussian control \cite{1100265, whittle1981risk} derives a policy that, unlike LQG, is dependent on the severity of resulting uncertainty in the system dynamics.
This risk sensitivity is determined by a scaling parameter $\sigma{\in}\mathbb{R}$ in a transformed objective,
\begin{align}
	-
	\frac{1}{\sigma}
	\log
	{\mathbb{E}}
	\left[
	\exp
	\left(
	\!-
	\sigma
	\left[
	C_T(\vx_T){+}\sum_{t=0}^{T-1}C_t(\vx_t,\vu_t)\right]
	\right)
	\right],\!
	\label{eq:risk}
\end{align}
which results in the adjusted Bellman equation,
\begin{align}
	\hspace{-0.5em}
	Q^\sigma_t(\vx_t, \vu_t)\! 
	&=\!
	C_t(\vx_t, \vu_t)\!
	+\!
	\textstyle\frac{1}{\sigma}
	\log
	{\mathbb{E}}\!
	\left[
	\exp\!
	\left(
	\!-
	\sigma
	V_{t+1}(\vx_{t+1})
	\right)
	\right].\!\notag
\end{align}
The expectation results in a transformed value function $V^\sigma$
\begin{align}
	\exp(-\sigma
	V^\sigma(\vx))
	&=
	\int\!
	\exp\!
	\left(
	\!-
	\sigma
	V(\vx)
	\right)
	\gN(\vx;\vmu_{\vx}, \sigEta{~}\!)\,
	\rd\vx.\notag
\end{align}
For a quadratic model of $V$,
this expectation is equivalent to adding two Gaussian random variables
${\mV_t^\sigma{\,=\,}(\sigEta{t}{\,+\,}\textstyle\frac{1}{\sigma}\mV_t\inv)\inv}$.
The $Q$-function naturally has a similar adjustment,
	$\mQ_t^\sigma \!=\! 
	\mC_t + \mF_{t+1}\tran\mV_t^\sigma\mF_{t+1}.$
As a result, the risk-sensitive optimal policy depends on the disturbance covariance.
For $\sigma{\,>\,}0$, the behavior is `risk seeking'.
For $\sigma{\,<\,}0$, the behavior is `risk averse'.
As $\sigma{\,\rightarrow\,}0$, we recover the nominal behavior, referred to as risk neutral.
For a nonlinear system, an approximate optimal solution is optimized through linearization \cite{farshidian2015risk}.

\subsection{Covariance Control}
\label{sec:cov_control}
While stochastic optimal control is typically concerned with optimizing the expected cost, methods have also been devised for controlling the state \emph{distribution}.
Covariance control \cite {4048352} specifically looks at constraining the mean and covariance of the terminal state distribution to a target $p(\vx_T^*)$.
The linear Gaussian setting has been extensively studied, for both discrete \cite{8264189} and continuous time \cite{7160692}, where it can be shown that a solution exists should the system be controllable and $\sigm{\vx}{*}{T}{-}\sigEta{T}{	\succeq}0$, given process noise covariance $\mSigma_{\veta_t}$.
The hard constraint can be tackled by decomposing the problem into feedforward control for the mean, and linear feedback control for the covariance~\cite{8264189}.
The discrete-time case has correspondences to relative entropy minimization and minimum-energy LQG \cite{532893, 8264189}, where the terminal cost corresponds to the Lagrange multiplier of the constraint.
The nonlinear Gaussian case has been tackled using stochastic differential dynamic programming \cite{yi2019nonlinear} and through the combination of sequential convex programs and statistical linearization~\cite{9147505}.
The problem can also be viewed as a form of optimal transport and the Schr{\"o}dinger bridge, which seeks to find the mapping, i.e. dynamical system, that transforms one distribution into another \cite{chen2016modeling}.

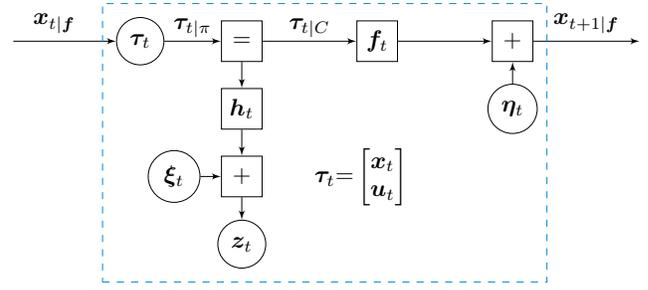
\begin{figure}[tb]
	\vspace{2em}
	\begin{tikzpicture}
[%
 node distance=20mm,
 auto,>=latex',
 box/.style={draw, minimum size=0.6cm},
 short/.style={node distance=10mm},
 transform canvas={scale=0.9} %
]
\node [yshift=0cm, xshift=0cm] (start) {}; %
\node[circle, draw, scale=1.0, right of=start] (joint) {$\vtau_t$} edge[<-] 
node[above,pos=0.6] {{$\vx_{t|\vf}$}}(start);
\node[short, below of=joint] (joint_vec) 
{};
\node[box, draw, right of=joint, xshift=-0.5cm] (x_fuse) {$=$} edge[<-] node[above=-2pt,pos=0.5] {{$\vtau_{t|\pi}$}} (joint);
\node[box, draw, below of=x_fuse, yshift=1cm] (g) {$\vh_t$} edge[<-](x_fuse);
\node[box, draw, below of=g, yshift=1cm] (xi_add)  {$+$} edge[<-] (g);
\node[circle, draw, left of=xi_add, xshift=1cm] (xi) {$\bm{\xi}_t$} edge[->](xi_add);
\node[short, right of=xi_add] (sig_xi)
{\hspace{14mm}
	$\vtau_t{=}
	\begin{bmatrix}
		\vx_t \\
		\vu_t
	\end{bmatrix}$}
;

\node[circle, draw, below of=xi_add, yshift=1cm] (z) {\hspace{0mm}$\vz_t$} edge[<-] node {} (xi_add);
\node[box, draw, right of=x_fuse, xshift=0cm] (f) {$\vf_t$} edge[<-] node[above=-2pt,pos=0.5] {{$\vtau_{t|C}$}}  (x_fuse); %
\node[box, draw, right of=f,xshift=0.0cm] (eta_add) {$+$} edge[<-]
node[above=-2pt,pos=0.5] {} (f);
\node[circle, draw, below of=eta_add, yshift=1cm] (eta) {$\veta_t$} edge[->] (eta_add);
\node[short, left of=eta] (sig_eta)
{}
;
\node[right of=eta_add] (end) {} edge[<-] node[above] {$\vx_{t+1|\vf}$}
  (eta_add); %

\node[box, subsectioncolor, dashed,inner sep=2mm,fit=(joint)(f)(g)(x_fuse)(xi_add)(z)(eta_add)] {};
\end{tikzpicture}
	\vspace{9.5em}
	\caption{Probabilistic graphical model of \itwoc for one timestep, following the notation of Loeliger et al. \cite{loeliger2007factor}, illustrating how incorporating the quadratic cost structure yields a Gaussian state-space model for inference.}
	\label{fig:factor_graph}
\end{figure}

\begin{figure*}[b]
	\begin{figeq}{joint distribution of the probabilistic graphical model for optimal control}{}
		{
			\vspace{-0.5\abovedisplayskip}
			\begin{equation}
				p(\mX_1^T, \mU_1^T, \mZ_1^T, \alpha)=
				p(\vx_1)
				\underbrace{p(\vz_T|\vx_T,\alpha)}_{\text{Terminal cost}}
				\textstyle\prod_{t=1}^{T-1}
				\underbrace{p(\vx_{t+1}|\vx_{t},\vu_t)}_{\text{Dynamics}}
				\underbrace{p(\vz_t|\vx_t,\vu_t,\alpha)}_{\text{Cost}}
				\underbrace{p(\vu_t|\vx_t)}_{\text{Controller}}
				\label{eq:i2cmodel}
			\end{equation}
		}
		\vspace{-2\belowdisplayskip}
	\end{figeq}
\end{figure*}

\section{Input Inference for Control}
\label{sec:i2c}
While progress has been made with respect to optimal control of systems subject to stochastic disturbances, progress is still hampered by the resulting required computations.
Thus we now use the perspective of optimal control as input estimation to derive an inference procedure for optimal control which estimates the optimal state-action distribution.

\subsection{Cost Functions and Constraints as Likelihoods}
\label{sec:likelihood}
Control-as-inference techniques require a \emph{belief in optimality} to perform optimal control\cite{Dayan97}, in order to construct a likelihood objective.
Broadly, one can consider a binary random variable $\mathcal{O}{\,\in\,}\{0,1\}$, for which $1$ indicates optimality~\cite{levine2018reinforcement}.
To ensure that a log-likelihood objective represents the cost, the exponential utility transform of the cost constructs a convenient likelihood \cite{Peters_PICML_2007}.
The resulting density is a Boltzmann distribution and introduces an inverse `temperature' $\alpha$ that scales the sharpness of this likelihood relative to the cost,
\begin{align}
	p(\mathcal{O}_t{=}1|\vx_t,\vu_t) &\propto \exp(-\alpha\, C_t(\vx_t, \vu_t)). \label{eq:cost_exponential}
\end{align}
As a result, the negative log-likelihood is an affine transform of the cost, which preserves convexity.
In this work, we exploit the fact that a quadratic cost function\footnote{While we assume explicit quadratic structure here, a local quadratic approximation of an arbitrary cost function could also be used like in differential dynamic programming, i.e. the Laplace approximation of \Eqref{eq:cost_exponential} from the inference perspective \cite{barberBRML2011}.}
defined in a space $\vz \in \mathbb{R}^{d_z}$ converts this distribution into a multivariate Normal, allowing approximate Gaussian inference methods to be used.
We parameterize the quadratic cost as a distance from target state $\vz_t^*$ with weight $\mTheta_t$, where space $\vz$ is defined through a transform $\vh_t(\vx,\vu)$ of the state-action variables
\begin{align}
	C_t(\vx_t,\vu_t) &= \mahalanobis{\vz_t^*-\vh_t(\vx_t,\vu_t)}{\mTheta_t}, \label{eq:l2_cost} \\ 
	p(\mathcal{O}_t{=}1|\vx_t,\vu_t) &= \gN(\vz_t{=}\vz_t^*|\vx_t,\vu_t,\mTheta_t,\alpha).
\end{align}
When viewed as a state-space model, this is equivalent to a likelihood with a `measurement' $\vz_t^*$ and `observation' model
\begin{align}
	\vz_t &= \vh_t(\vx_t,\vu_t) + \vxi_t,\;\vxi_t \sim \gN[\vzero, \alpha\mTheta_t].
	\label{eq:cost_gaussian}
\end{align}
This motivates the use of Bayesian smoothing \cite{simo_filt} to perform inference on this state-space model, despite considering a fully-observed problem, as the cost acts as an observation which guides the latent state-action trajectory towards optimality.
This modeling assumption is motivated not only by the popularity of quadratic costs in optimal control; the resulting Gaussian cost likelihood yields a state-space model (Figure \ref{fig:factor_graph}) that defines a finite-horizon inference problem (\Eqref{eq:i2cmodel}) that can be tackled using approximate Gaussian inference, for which effective methods of analysis and computation are available \cite{loeliger2007factor,simo_filt}.
This inference problem evokes the use of the expectation maximization algorithm for dynamical system estimation \cite{shumway1982approach,SCHON201139}, but with the addition of input estimation \cite{bruderer2015input}.
In this case, the expectation maximization (EM) algorithm (Algorithm \ref{algo:i2c}) consists of an E-step estimating the optimal state distribution given the cost and priors, and an M-step updating the priors and re-estimating the cost temperature $\alpha$. 
This contrasts with prior work in which $\alpha$ is a constant hyperparameter, so only a single (but iterated) E-step is performed \cite{rawlik2013probabilistic}.
The perspective of input estimation, treating $\vu$ as a latent variable, provides a flexibility during inference that allows for both closed- and open-loop optimization and adaptive exploration.
Therefore, we refer to this method as input inference for control (\itwoc).

\begin{algorithm}[tb]
	\SetAlgoLined
	\KwIn{Probabilistic graphical model $p(\mZ_1^T,\mX_1^T,\mU_1^{T-1}, \alpha_0)$, iterations $N$}
	\KwOut{Posterior controllers, $p(\vu_t\mid\vx_t, \vz_{1:T})$}
	\For{$i \gets 1$ to $N$}{
		E-step: Bayesian smoothing to infer $p(\mX_1^T, \mU_1^{T-1}|\mZ_1^T)$ given $\alpha_{i-1}$, (Section \ref{sec:slvm})\\
		M-step: Update $\alpha_i$ and priors given $p(\mX_1^T, \mU_1^{T-1}|\mZ_1^T)$
	}
\caption{Input inference for control} \label{algo:i2c}
\end{algorithm}

The EM approach uses Bayes' rule to iteratively weigh the prior controller against the likelihood (i.e. control cost) to compute the posterior state-action distribution, which is the prior for the next iteration, as the dynamics and $p(\vx_1)$ are fixed.
This procedure is known as posterior policy iteration~\cite{rawlik2013probabilistic},
\begin{align*}
	p_{i+1}(\mX,\mU) &{\,=\,} p_{i}(\mX,\!\mU|\mZ)
	{\,\propto\,} p_i(\mX,\!\mU)\!\exp(\!-\alpha\textstyle\sum_t\! C_t(\vx_t,\!\vu_t)),\\
	p_{i+1}(\mU|\mX)
	&{\,\propto\,} p_i(\mU|\mX) \exp(\!-\alpha\textstyle\sum_t\! C_t(\vx_t,\!\vu_t)). 
\end{align*}
The success of this procedure rests on three aspects: An initial prior $p_0(\mU|\mX)$ that provides sufficient exploration, accurate inference to evaluate the controller, and an $\alpha$ that adequately calibrates the likelihood to effectively optimize under uncertainty by regularizing against greedy optimization.

\subsection{Inference of Sequential Latent Variables}
\label{sec:slvm}
Factorizing the joint distribution into its Markovian structure (\Eqref{eq:i2cmodel}) defines a graphical model that can be efficiently solved using recursive Bayesian inference over time.

\begin{definition}{general \itwoc inference}{generalitwoc}
	{
	\vspace{-1em}
	\begin{align*}
	\intertext{\hspace{-1em}Control, where $p(\vu_t|\vx_t)$ is a stochastic control law $\vpi_t(\vx)$}
		\hspace{-6em}p(\vu_t | \vx_t, \vz_{1:t{-}1}) &\propto  p(\vu_t|\vx_t)p(\vx_{t} | \vz_{1:t{-}1}) %
	\intertext{\hspace{-1em}Forward optimization (innovation from the cost)}
		p(\vx_{t}, \vu_t | \vz_{1:t}) \!&\propto\! p(\vz_{t}|\vx_t,\!\vu_t)
		p(\vx_t, \vu_t|\vx_t,\vz_{1:t{-}1}\!)
	\intertext{\hspace{-1em}Dynamics prediction}
		p(\vx_{t+1}|\vz_{1:t}) &= 
		{\int} p(\vx_{t+1} | \vx_t, \vu_t)
		p(\vx_t, \vu_t | \vz_{1:t})\rd\vx_t\rd\vu_t
	\intertext{\hspace{-1em}Backward optimization (smoothing)}
		p(\vx_{t}, \vu_{t}|\vz_{1:T}) &= \notag\\
		&\hspace{-4.5em}
		p(\vx_{t}, \vu_{t} | \vz_{1:t})
		\!\int\!\left[\frac{p(\vx_{t+1} | \vx_t, \vu_t)p(\vx_{t+1} |\vz_{1:T})}{p(\vx_{t+1} | \vz_{1:t})}\right] \rd\vx_{t+1} %
	\end{align*}
	\vspace{-1em}
	}
\end{definition}

Note that \itwoc inference mirrors trajectory optimization, as both have a forward pass that simulates the current controller and a backward pass that acts to improve the trajectory.
These steps are presented in Definition \ref{def:generalitwoc}.
While the equations are equivalent to Bayesian smoothing, the control step is unique to \itwoc, due to the latent variable being the joint state-action distribution.

\subsection{The Gaussian Assumption}
\label{sec:gaussian}
For tractability and convenience, we assume a Gaussian state-action distribution for the remainder of this article.
As the central SOC problem (Equation 1) assumes a unique optimal trajectory, with approximate inference we use the Gaussian density to approximate the posterior about the maximum a posterori solution.
This assumption is supported by the Bernstein-von Mises theorem\cite{vaart_1998}, which states a posterior will converge to a multivariate Normal about the maximum likelihood solution under certain regularity conditions. 
From the physical perspective, the Gaussian assumption of the dynamics disturbances and initial state distribution is supported by the central limit theorem\cite{vaart_1998}.
The subsequent assumption of a Gaussian state-action trajectory is supported by adopting Gaussian controllers and the assumption that this dynamical system is adequately controlled, thus maintaining unimodality in the state distribution.
Therefore, Gaussian \itwoc is limited to solving for the maximum a posterori solution, and is unable to represent a more sophisticated controller than can reason over multimodal optimal trajectories.
However, Gaussian filtering and smoothing has several attractive numerical qualities, namely exact inference in the linear setting, and well-developed approximate message passing techniques in the nonlinear setting \cite{loeliger2007factor,8450699,simo_filt}.
Finally, Gaussians are convenient to marginalize and condition on, such that $p(\vu|\vx)$ results in time-varying linear (Gaussian) controllers.
Therefore, while the Gaussian assumption is limiting, its locally linear quadratic approximations can be compared to linear quadratic optimal control approaches, which have been widely adopted.

\begin{figure}[!tb]
\begin{definition}{gaussian \itwoc inference}{gaussianitwoc}
	{
		Control, where $\vpi_t(\vx)=\gN(\mK_t^{\vpi}\vx + \vk_t^{\vpi},\mSigma^\vpi_{t}$)
		\begin{align}
			\mK_t^{\vpi} &:= \mSigma_{\vu\vx_t}\mSigma_{\vx_t}\inv\label{eq:i2c_gaussian_K} \\ 
			\vk_t^{\vpi} &:= \vmu_{\vu_t}-\mK_t^{\vpi}\vmu_{\vx_t}\notag\\
			\mSigma^\vpi_{t} &:= \mSigma_{\vu_t}-\mSigma_{\vu\vx_t}\mSigma_{\vx_t}\inv\mSigma_{\vu\vx_t}\tran\label{eq:i2c_covariance_control}\\
			\vmu_{\vu_{t|\vpi}} &= \mK^\vpi_t\vmu_{\vx_{t|\vf}} + \vk^\vpi_t\notag\\
			\mSigma_{\vu_{t|\vpi}} &= \mSigma^\vpi_{t} + \mK^\vpi_t \mSigma_{\vx_{t|\vf}}{\mK_t^\vpi}\tran\notag
		\end{align}
		Forward optimization (innovation from the cost)
		\begin{align}
			p(\vtau_{t|\vpi}) &= p(\vx_t, \vu_t | \vz_{1:t-1}) = \notag\\
			&\hspace{-1em}
			\gN\!\left(
			\begin{bmatrix}
				\vmu_{\vx_{t|\vf}} \\
				\vmu_{\vu_{t|\vpi}} \\
			\end{bmatrix}\!,\!
			\begin{bmatrix}
				\mSigma_{\vx_{t|\vf}}             & \hspace{-1em}\mSigma_{\vx_{t|\vf}}{\mK_t^{\vpi}}\tran\\
				\mK_t^{\vpi}\mSigma_{\vx_{t|\vf}} &\hspace{-1em}\mSigma_{\vu_{t|\vpi}}\\
			\end{bmatrix}
			\right)\!\!\notag\\
			\vz_{t|\vpi} &= \vh_t(\vtau_{t|\vpi}) + \vxi_{t}\notag\\
			p(\vz_{t|\vpi},\vtau_{t|\vpi}) &=\notag\\
			&\hspace{-1em}\gN\!\left(
			\begin{bmatrix}
				\vmu_{\vz_{t|\vpi}} \\
				\vmu_{\vtau_{t|\vpi}} \\
			\end{bmatrix}\!,\!
			\begin{bmatrix}
				\mSigma_{\vz_{t|\vpi}}               &\hspace{-1em}\mSigma\tran_{\vtau_{t|\vpi}\vz_{t|\vpi}} \\
				\mSigma_{\vtau_{t|\vpi}\vz_{t|\vpi}} &\hspace{-1em}\mSigma_{\vtau_{t|\vpi}}\\
			\end{bmatrix}
			\right) \label{eq:observation_joint} \\ 
			\mK_t^C &:= \mSigma_{\vtau_{t|\vpi}\vz_{t|\vpi}}\mSigma_{\vz_{t|\vpi}}\inv\text{ (Kalman gain),}\notag\\
			\vmu_{\vtau_{t|C}} &= \vmu_{\vtau_{t|\vpi}} + \mK_t^C(\vz_t - \vmu_{\vz_{t|\vpi}})\notag\\
			\mSigma_{\vtau_{t|C}} \notag
			&
			\!=\!\mSigma_{\vtau_{t|\vpi}}\!-\!\mK_t^C\mSigma_{\vz_{t|\vpi}}{\mK_t^C}\tran\notag
			\intertext{Dynamics prediction, where $\vx_{t+1|\vf} = \vf_t(\vtau_t) + \veta_{t}$}
			\hspace{-1em}p(\vx_{t+1|\vf}, \vtau_t) &=\notag\\
			&\hspace{-3em}\gN\!\left(
			\begin{bmatrix}
				\vmu_{\vx_{t+1|\vf}} \\
				\vmu_{\vtau_{t|C}} \\
			\end{bmatrix}\!,\!
			\begin{bmatrix}
				\mSigma_{\vx_{t+1|\vf}}                    &\hspace{-1.2em}\mSigma\tran_{\vx_{t+1|\vf}\vtau_{t|C}} \\
				\mSigma_{\vx_{t+1|\vf}\vtau_{t|C}} &\hspace{-1.2em}\mSigma_{\vtau_{t|C}}\\
			\end{bmatrix}
			\right)\!\label{eq:dynamics_joint}
		\end{align}
		Backward optimization (smoothing)
		\begin{align}
			\mK_t^{s} &:= \mSigma_{\vx_{t+1|\vf}\vtau_{t|C}}\tran\mSigma_{\vx_{t+1|\vf}}\inv\notag\\
			\vmu_{\vtau_t|T} &= \vmu_{\vtau_t|C} + \mK_t^{s}(\vmu_{\vx_{t+1|T}}-\vmu_{\vx_{t+1|\vf}})\notag\\
			\mSigma_{\vtau_t|T} &= 
			\mSigma_{\vtau_t|C}{+} \mK_t^{s}(\mSigma_{\vx_{t+1|T}}{-}\mSigma_{\vx_{t+1}|\vf}){\mK_t^{s}}^\top \label{eq:i2c_gaussian_end}
		\end{align}
		\vspace{-2em}
	}
\end{definition}
\end{figure}

Definition \ref{def:gaussianitwoc} describes the explicit message passing for Gaussian \itwoc from Definition \ref{def:generalitwoc}.
To compactly delineate each step, we use $t|\vpi$, $t|C$ and $t|\vf$ to indicate a state after control, cost innovation and dynamics respectively.
See Figure~\ref{fig:factor_graph} for a factor graph visualization of the messages.
As with standard Bayesian smoothing notation, $t$ and $t|T$ describes the prior and posterior distribution. 
The initial state distribution is $\vx_{1|\vf}{\,\sim\,}p(\vx_1)$.
Note, the computation of the joint distributions (Equations \ref{eq:observation_joint} and \ref{eq:dynamics_joint}) depends on the chosen message passing method, which is approximate in the nonlinear setting (Section \ref{sec:nonlinear}).
While Equations \ref{eq:i2c_gaussian_K}-\ref{eq:i2c_gaussian_end} are familiar to those experienced in Gaussian state estimation, it is not immediately clear from the expressions that this computation performs optimal control.
To reveal this connection, we must consider the linear Gaussian setting and its duality to linear quadratic optimal control.

\subsection{Linear Gaussian Inference \& Linear Quadratic Control}
\label{sec:linear_gaussian}
In the the case when the dynamics $\vf$ and the cost transform $\vh$ are affine transformations, the Gaussian message passing described in Section \ref{sec:gaussian} can be performed exactly.
Specifically, for notational simplicity we consider
	$C_t(\vtau){\,=\,}\mahalanobis{\vtau^*_t{\,-\,}\vtau}{\mC_t}$,
	and 
	$\vx_{t+1}{\,=\,}\mF_t \tau_t{\,+\,}\bar{\vf}_t{\,+\,}\veta_t$
as in Section \ref{sec:soc}.

Exact message passing allows us to examine the inference computation in closer detail, and therefore concretely derive the correspondence to optimal control.
However, in order to do this, we must first carefully distinguish the prior, likelihood and posterior, where the prior is the filtered distribution and the posterior is the smoothed distribution
\footnote{Here, the prior at timestep $t$ refers to the belief after inference forward in time for $1{:}t$ (i.e. filtering), while the posterior refers to the belief after inference over the whole trajectory $1{:}T$, which incorporates smoothing.}.
Adopting notation from previous work \cite{loeliger2007factor,i2corl}, we use $\rightarrow$ for the prior and $\leftarrow$ for the likelihood, so the posterior $p(\vx)\!\propto\!p(\overrightarrow{\vx})p(\overleftarrow{\vx})$, where
\begin{align}
	\mSigma &= 
	(\overrightarrow{\mLambda} + \overleftarrow{\mLambda})\inv = 
	\overrightarrow{\mSigma} - \overrightarrow{\mSigma}(\overrightarrow{\mSigma} + \overleftarrow{\mSigma})\inv\overrightarrow{\mSigma}.\label{eq:woodbury_cov}
\end{align}
Clearly recovering the optimal control expressions from the messages is challenging due to the nested regularization provided by the probabilistic computation.
We use the relation between the $Q$-function and state-action log-likelihood as a compact alternative.
To demonstrate this control equivalence, in the following passage we consider the limit where the prior becomes `uninformative', that is  $p(\vx){\rightarrow}p(\overleftarrow{\vx})$, and the dynamics deterministic ($\mSigma_{\veta_t}{\,\rightarrow\,}\vzero$).
Numerically, uninformative priors occur when the relative uncertainty in the prior is significantly larger than the likelihood. 
We use $\rightarrow$ in the following expressions to express this limiting case.
Revisiting the smoothing step (\ref{eq:i2c_gaussian_end}),
using Woodbury's inversion lemma \cite{petersen2008matrix}, we can express the posterior using \Eqref{eq:woodbury_cov} as
\begin{align}
	\mSigma_{\vx_{t+1|T}} &=\! \mfSigma_{\vx_{t+1|\vf}}{-}\mfSigma_{\vx_{t+1|\vf}}(\mfSigma_{\vx_{t+1|\vf}}{+}\mbSigma_{\vx_{t+1}})\inv\mfSigma_{\vx_{t+1|\vf}}.\notag\\
\intertext{Substituing into the smoothing update (Equation \ref{eq:i2c_gaussian_end}),}
	\mSigma_{\vtau_t|T} &= 
\mfSigma_{\vtau_t|C}{-} \mfSigma_{\vtau_t|C}\mF_t\tran(\mfSigma_{\vx_{t+1|\vf}}+\mbSigma_{\vx_{t+1}})\mF_t\mfSigma_{\vtau_t|C},\notag
\end{align}
given $\mK_t^s{\,=\,}\mSigma_{\vtau_t|C}\mF_t\tran\mSigma_{\vx_{t+1|\vf}}\inv$ in the linear Gaussian setting.
For affine dynamics, $\mSigma_{\vx_{t+1|\vf}} = \mF_t\mSigma_{\vtau_t|C}\mF_t\tran + \sigEta{t}$, so we can use a matrix inversion identity\footnote{\label{note1}$\mA\inv - \mA\inv\mU(\mV\mA\inv\mU + \mB\inv)\inv\mV\mA\inv = (\mA + \mU\mB\mV)\inv$}\cite{petersen2008matrix} for $\mLambda_{\vtau_t|T}$, where
\begin{align}
	\mLambda_{\vtau_t|T} &=
	\mfLambda_{\vtau_t|C} + \mF_t\tran(\sigEta{t}+\mbSigma_{\vx_{t+1}})\inv\mF_t.\label{eq:pen_bellman}
\end{align}
The future state posterior can also be expressed as $\mSigma_{\vx_{t+1|T}}{=}(\mfSigma_{\vx_{t+1|\vf}}\inv{\,+\,}\mbSigma_{\vx_{t+1}}\inv)\inv$, and given the uninformative prior $\mLambda_{\vx_{t+1|T}}{\rightarrow}\mbLambda_{\vx_{t+1}}$.
Moreover, from the innovation step during filtering $\mLambda_{\vtau_t|C}{\,=\,}\mLambda_{\vtau_t|\vpi}{\,+\,}\alpha\mC_t$, and $\sigEta{t}$ is negligible for deterministic dynamics, \Eqref{eq:pen_bellman} reduces to
\begin{align}
	\mLambda_{\vtau_t|T} &\rightarrow \alpha\mC_t + \mF_t\tran\mLambda_{\vx_{t+1|T}}\mF_t,\label{eq:i2c_to_bellman}
\end{align}
since $\mfLambda_{\vtau_t|\vpi}$ is uninformative.
We can now identify \Eqref{eq:i2c_to_bellman} as the $Q$ update from \Eqref{eq:Q_update}, with $\mLambda_{\vtau_t|T}{\,\equiv\,}\mQ_t$ and $\mLambda_{\vx_{t+1|T}}{\,\equiv\,}\mV_t$.
Reflecting on the deterministic dynamics assumption where $\sigEta{t}$ is negligible, the reader may recognize that the $(\sigEta{t}{+}\mbSigma_{\vx_{t+1}})\inv{\,\rightarrow\,}\mLambda_{\vx_{t+1|T}}$ step is the same regularization introduced to the value function in the risk-sensitive formulation in Section \ref{sec:risk},
${\mV_t^\sigma{\,=\,}(\sigEta{t}{\,+\,}\textstyle\frac{1}{\sigma}\mV_t\inv)\inv}$.
We discuss this correspondence in more detail in Section \ref{sec:i2c_risk}.

Having shown that the Gaussian posteriors relate to $Q$ and $V$, we now illustrate the maximization step of dynamic programming through marginalization.
For brevity, we drop the time index and conditioning indicator.
To relate the conditional of the $\vtau$ precision to the optimal linear feedback law, consider the block-wise inversion of the joint covariance 
\begin{align}
	\hspace{-.5em}\mLambda_{\vtau}&{=} 
	\begin{bmatrix}
		\mSigma_{\vx}      & \mSigma_{\vu\vx}\tran\\
		\mSigma_{\vu\vx}   & \mSigma_{\vu} \\
	\end{bmatrix}\inv
	=
	\begin{bmatrix}
		\mLambda_{\vx\vx}      & \mLambda_{\vu\vx}\tran\\
		\mLambda_{\vu\vx} & \mLambda_{\vu\vu} \\
	\end{bmatrix}\notag\\
	&{=}
	\begin{bmatrix}
	\mSigma_{\vx}\inv{+}
	\mSigma_{\vx}\inv
	\mSigma_{\vu\vx}
	\mSigma_{*}\inv
	\mSigma_{\vu\vx}
	\mSigma_{\vx}\inv
	&\hspace{-0.7em}-\mSigma_{\vx}\inv\mSigma_{\vu\vx}\tran\mSigma_{*}\inv\\
	-\mSigma_{*}\inv\mSigma_{\vu\vx}\mSigma_{\vx}\inv & \mSigma_{*}\inv \\
	\end{bmatrix}\!\!\label{eq:precision_joint}
\end{align}
where $\mSigma_{*} = \mSigma_{\vu} {-} \mSigma_{\vu\vx}\mSigma_{\vx}\inv\mSigma_{\vu\vx}\tran$ (Schur complement), so
\begin{align}
-\mLambda_{\vu\vu}\inv\mLambda_{\vu\vx} &\equiv -\mQ_{\vu\vu}\inv\mQ_{\vu\vx}, \notag\\
 &= 
\mSigma_{*}\mSigma_{*}\inv\mSigma_{\vu\vx}\mSigma_{\vx}\inv =
\mSigma_{\vu\vx}\mSigma_{\vx}\inv. \label{eq:controller_Q_i2c_relation}
\end{align}
Therefore, \Eqref{eq:controller_Q_i2c_relation} relates the $Q$ function-derived controller from \Eqref{eq:optimal_control} to the \itwoc controller in \Eqref{eq:i2c_gaussian_K}.

Finally, we use the joint precision in \Eqref{eq:precision_joint} again to derive the value function update from \Eqref{eq:V_update}. 
The value function corresponds to the marginal of $\vx$ from the joint.
To understand its connection to the $Q$ function, we consider the how the joint precision term $\mLambda_{\vx\vx}$ in \Eqref{eq:precision_joint} relates to the \emph{marginal}  state precision $\mSigma_{\vx}\inv{\,=\,}\mLambda_{\vx}$,
\begin{align}
	\mLambda_{\vx\vx} &\!=\! \mSigma_{\vx}\inv \!+\!
	\mSigma_{\vx}\inv
	\mSigma_{\vu\vx}
	\mSigma_{*}\inv
	\mSigma_{\vu\vx}
	\mSigma_{\vx}\inv\\
	&=
	\mLambda_{\vx} +
	\mLambda_{\vx\vu}
	\mLambda_{\vu\vu}\inv
	\mLambda_{\vu\vx},\notag\\
\therefore \mLambda_{\vx} &= \mLambda_{\vx\vx} - \mLambda_{\vx\vu}
\mLambda_{\vu\vu}\inv
\mLambda_{\vu\vx},
\end{align}
which corresponds to the $V$ function update in \Eqref{eq:V_update}.
As the dynamic programming method maintains a separate value and $Q$ function, the value function is updated according to the equations in \Eqref{eq:V_update}.
However, through manipulating the Gaussian distributions we can simply marginalize $\vx$ from $\vtau$ and compute $\mLambda_{\vx}$ directly, and by working with the joint distribution throughout we avoid computing $Q$ and $V$ separately.

This analysis is closely related to the `soft' Bellman backup from reinforcement learning \cite{levine2018reinforcement}, and the insight that a sample covariance approximates the inverse Hessian of its log-likelihood under the Gaussian assumption used in sampled differential dynamic programming \cite{saDDP1}.

\subsection{Approximate Nonlinear Gaussian Inference}
\label{sec:nonlinear}

We now review methods for Gaussian message passing when the exact linear rules are not possible, as this is important for many interesting problems.
To do so, we revisit the limiting assumption in \Eqref{eq:i2c_to_bellman},
and replace $\vh_t(\vtau_t)$ with a locally linear approximation $\mH_t\vtau_t + \bar{\vh}_t$, which may be obtained via function linearization or `statistical linearization' using the quadrature or sample estimates.
Using this linearized model, our equivalent cost coefficients should be transformed for optimizing $\vtau$, i.e. $\mC^\vtau_t{\,=\,}\mH\tran\mC_t\mH$.
Starting from the simplified Kalman update of $\mSigma_{\vtau_t|C}$, 
using the matrix inverse identity$^{\ref{note1}}$, we derive
\begin{align}
	\mSigma_{\vtau_t|C} &= \mSigma_{\vtau_t|\vpi}{-} \mSigma_{\vtau_t|\vpi}\mH_t\tran(\mH\mSigma_{\vtau_t|\vpi}\mH_t\tran{+} \textstyle\frac{1}{\alpha}\mC_\vtau\inv)\inv\mH_t\mSigma_{\vtau_t|\vpi}\notag\\
	\mLambda_{\vtau_t|C} &= \mLambda_{\vtau_t|\vpi} + 
	\alpha\mH_t\tran\mC_\vtau\mH_t,\text{ as required.}\notag
\end{align}

Various approximation techniques for nonlinear Gaussian inference are available, compare Figure \ref{fig:inference}:

\subsubsection{Linearization}
\label{subsec:linearization}
Approximating the dynamics with a local linearization is attractive as it allows the linear Gaussian message passing rules to be adopted.
Methods that adopt this approximation, such as the extended Kalman filter, are popular due to this convenience \cite{anderson2012optimal}.
Like with differential dynamic programming, a second-order approximation can also be used, though this is costly.
Local linearization approximations are limited in their accuracy, not only for highly nonlinear systems, but also for large state uncertainties.
They typically require greater regularization or iterative computation to refine the linearization point.
Moreover, while Jacobians can be straightforward to calculate, due to advancements in automatic differentiation, they are still $\mF_t \in \mathbb{R}^{d_x \times d_\tau}$ objects and therefore expensive to evaluate without optimized implementations.
Linearization is also brittle numerically when applied to discontinuities such as constraints.
Also, as with iLQR, EM with the linearization-based inference corresponds to Gauss-Newton optimization~\cite{bell1994iterated}.

\subsubsection{Spherical Cubature Quadrature}
\label{subsec:cubature}
Quadrature rules construct evaluation points and weights in order to evaluate integrals for specific functions.
For Gaussian densities, the 3rd order spherical cubature rule is popular \cite{arasaratnam2009cubature}, as it requires $2d$ points, scaling linearly with dimensionality $d$.
This rule is also a special case of the unscented transform, popularized by the unscented Kalman filter \cite{ukf}.
The cubature quadrature rules are derived by computing `sigma' points about the mean in each axis via the Cholesky decomposition of the covariance.
The unscented transform includes the mean, and therefore requires $2d{+}1$ points, as well as two additional hyperparameters.
While the unscented transform may provide an additional performance boost with hyperparameter tuning, we omit evaluation here and focus on the spherical cubature rules.
Moreover, this rule relates to linearization through implicitly approximating the Jacobian via finite difference using the sigma points \cite{schei1997finite}.

\begin{figure}[!t]
	\input{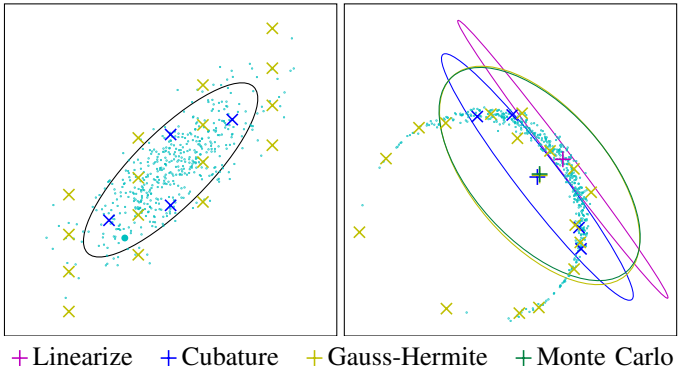}
	\vspace{-.5em}
	\begin{minipage}[t][0.8cm][t]{\columnwidth}
	\begin{tikzpicture}
	
	\begin{axis}[
		hide axis,
		width=8.5cm,
		xmin=10, xmax=50,
		ymin=0, ymax=1.0,
		legend cell align={center},
		legend columns=4,
		legend style={/tikz/every even column/.append style={column sep=0.3cm}, draw=none},
		]
		]
		
		\addlegendimage{semithick, magenta, mark=+, mark size=3, mark options={solid}, only marks}
		\addlegendentry{Linearize};
		\addlegendimage{semithick, blue, mark=+, mark size=3, mark options={solid}, only marks}
		\addlegendentry{Cubature};
		\addlegendimage{semithick, yellow, mark=+, mark size=3, only marks}
		\addlegendentry{Gauss-Hermite};
		\addlegendimage{semithick, green, mark=+, mark size=3, only marks}
		\addlegendentry{Monte Carlo};
	\end{axis}
	
\end{tikzpicture}
	\vspace{-1em}
	\end{minipage}
	\caption{An illustration of different approximate inference methods when propagating a bivariate Gaussian through a nonlinear function.
	The Monte Carlo samples (\ref{sample}) indicate the distribution becomes highly non-Gaussian, but evaluating these samples is computationally intensive.
	Linearizing the function returns an approximation that is highly localized around its (inaccurate) mean prediction.
	Cubature quadrature  (\ref{cubature_point}) uses $2d$ points, but improves the estimate, particularly in the mean.
	4th-degree Gauss-Hermite (\ref{gh_point}) uses $d^4$ points, but almost directly matches the Monte Carlo estimate in mean and covariance.}
	\label{fig:inference}
\end{figure}
 
\subsubsection{Gauss-Hermite Quadrature}
\label{subsec:mc}
While state estimation requires computationally efficient inference methods for real-time use, for `offline' trajectory optimization we are able to invest computational cost for greater inference accuracy.
There are many approaches for achieving this for Gaussian distributions.
For example, Monte Carlo sampling will converge, regardless of state dimension, due to the central limit theorem. 
However, for time-series inference sequential Monte Carlo methods are preferred, and in practice we did not find the Monte Carlo inference to be numerically stable for the desired planning horizons.
Continuing with the use of quadrature rules, Gauss-Hermite (G-H) rules \cite{855552} are designed integrals of the form $\int\exp(x^2)f(x)dx$, therefore appropriate for computing Gaussian moments.
These rules are practically limited in the multivariate setting, as extending the quadrature points to a mesh results in exponential increase in points, $d^p$ for state dimension $d$ and degree $p$. 
We use it here as a baseline for accurate inference with \itwoc{}.
The Gauss-Hermite message passing is the same as cubature quadrature, however the points and weights for a given degree must be obtained.
Moreover, in the multivariate settings, the (1D) points and weights must be transformed to the appropriate mesh structure \cite{simo_filt}. 

\subsection{Expectation Maximization for Control}
\label{sec:em_inference}
While the E-step returns an estimate of the optimal latent state-action distribution using Bayesian smoothing,
the inverse temperature of the cost likelihood remains unknown.
Prior approaches treat this term as a fixed constant \cite{toussaint2009robot,rawlik2013stochastic},
many adopt an adaptive approach, such as by satisfying an explicit KL constraint \cite{levine2013variational}.
We propose an alternative strategy where $\alpha$ is treated as an unknown model parameter, which we can optimize in the M-step of EM.
This natural adaptivity removes the need to tune both $\alpha$ and the control prior as hyperparameters, as they jointly influence the E-step, and provides a means of automatically scaling this term for an arbitrary cost function.
The question of which strategy is best ultimately rests on the underlying optimization problem.
If the problem is convex (e.g. LQG), $\alpha$ can be fixed to a large value to avoid unnecessary regularization.
For highly non-convex problems, this regularization is beneficial, but for weakly non-convex problems, tuning a constant $\alpha$ could outperform an adaptive strategy.
There are also numerous alternative adaptive strategies, such as a linear schedule.
In Section \ref{sec:optimization}, this adaptive strategy is motivated in the context of Gauss-Newton optimization heuristics.

For Gaussian \itwoc, this expected log-likelihood is convex in $\alpha$, which affords the closed-form update\cite{i2corl},
\begin{align}
	\alpha &= \frac{(T{-}1)d_z + d_{z_T}}
	{\sum_t\tr\{{\mTheta_t\E[\delta\vz_t\delta\vz_t\tran]}\}},\text{ where }\delta\vz_t = \vz^*_t - \vz_t.\label{eq:alpha_update}
\end{align}  
As $\tr\{{\mTheta_t\E[\delta\vz_t\delta\vz_t\tran]}\} {\,=\,} \E[\delta\vz_t\tran\mTheta_t\delta\vz_t]$, the expected cost, $\alpha$ is calibrated using the total trajectory cost averaged over the belief, time horizon and state dimensionality.
During the M-step, we also update the control priors with the posterior $p(\mX, \mU|\mZ)$ for the next iteration. 

In the nonlinear setting, this M-step is approximate as $\E[\delta\vz_t\delta\vz_t\tran]$ can not be computed exactly. 
Motivated by the idea of a regularized M-step, we restrict the update of $\alpha$ by applying a KL constraint to $\vxi$, which can be shown to be equivalent to a ratio constraint on the $\alpha$ update \cite{i2corl}.
As $\alpha$ is an inverse temperature, it controls the sharpness of the likelihood. 
For sufficiently stochastic problems, a sharp likelihood can lead to exploitation of the random dynamics disturbance, as the `signal-to-noise ratio' of the simulated control belief against the dynamics noise makes it difficult to optimize the controls under the stochastic dynamics.
This tendency to exploit uncertainty during optimization is coined `optimism'~\cite{levine2018reinforcement}. 
Failing to regularize $\alpha$ can result in overly optimistic optimization, which can exploit state uncertainty arising from both the dynamics or approximate inference.

The complete \itwoc algorithm is summarized in Algorithm~\ref{algo:i2c}.
The remaining hyperparameters to discuss are the control priors, specified for $\vu$.
These priors play a range of roles in the trajectory optimization and should be designed appropriately. 
Firstly they act as a learning rate, so lower variance priors will result in slower optimization.
Secondly, they represent a source of entropy, so the variance also dictates the amount of possible exploration during optimization. 
Another design consideration is the nonlinearity of the system and inference choice, as the prior should be chosen to maintain the validity of the Gaussian assumption.

Regarding convergence guarantees, EM achieves monotonic improvement when inference is exact, and will converge to a unique optimum when the likelihood is convex \cite{em_properties}.
When the E-step is approximate, likelihood optimality guarantees are limited by the inference accuracy \cite{JMLR:v6:gunawardana05a}.
This limitation motivates the adoption of Monte Carlo methods, as accuracy can be improved through a greater number of particles~\cite{SCHON201139}.

A discussion on the computational complexity of Gaussian \itwoc depends on the choice of inference method, but generally it is dominated by the matrix inversions of the largest covariance matrix.
As the covariance inversion can be framed as solving a linear system with a positive definite matrix, the complexity is less severe than explicit inversion.
Table \ref{tab:time_tab} provides an empirical study of computational complexity for a complex dynamical system.
It is also worth highlighting that the Bayesian smoothing approach adopted by \itwoc requires fewer inversions than the belief propagation approach proposed for PPI \cite{rawlik2013probabilistic}. 
For a single forward and backward pass on a linear system, \itwoc is $\gO(3d_x^3 + d_u^3)$ while PPI is $\gO(7d_x^3 + 3 d_u^3)$ if $d_z{\,=\,} d_x{\,+\,}d_u$.

\subsection{Posterior Policy Iteration}

Input inference for control can be viewed as an instantiations of approximate inference control \cite{toussaint2009robot} and posterior policy iteration \cite{rawlik2013stochastic, rawlik2013probabilistic}, as it estimates the posterior controller introduced in Section \ref{sec:likelihood} using approximate inference.
\itwoc is motivated as a cogent realization of this approach compared to the algorithmic definition of PPI (Algorithm 1, \cite{rawlik2013probabilistic})
It is motivated closer to approximate Bayesian smoothing, rather than a regularized Riccati solver, to aid algorithmic developments and insights.
The algorithmic realization of PPI fixes the prior controller, so only the posterior state trajectory is optimized using local linearization.
However, for many complex tasks, the posterior controller must be iteratively evaluated, e.g. due to actuator constraints.
Secondly, using posterior linearizations for the subsequent forward pass, rather than directly evaluating the dynamics, has severely limited accuracy in practice when performing nonlinear trajectory optimization.
Conversely, \itwoc considers the joint state-action distribution, so the dynamics and posterior controller are iteratively evaluated at each E-step.
This realization enjoys a simpler implementation, fewer hyperparameters and lower computational complexity.
This perspective, more grounded in approximate inference, aids translation into alternative inference techniques such as sequential Monte Carlo, and motivates connections to stochastic control, such as covariance control.

\section{Approximate Inference for Optimization}
\label{sec:optimization}
To clarify the difference between the expectation maximization proposed in Section \ref{sec:i2c} and the prior work that used iterative inference, we focus on the optimization perspective.
To do so, we examine the simpler, but closely related, setting of a nonlinear inverse problem, to find input $\vx$ that maps to target $\vy^*$ through $\vf: \sR^{d_x} \rightarrow \sR^{d_y}$, which is assumed differentiable,
$\min_\vx |\vy^* - \vf(\vx)|^2$.
As before, we can optimize this problem using Gaussian inference, by framing it as a latent variable model with an unknown $\alpha$, where the belief in $\vx$ is inferred iteratively
\begin{align*}
	\vy_i &= \vf(\vx_i) + \vxi_i, 
	\; \vx_i \sim \gN(\vmu_i, \mSigma_i),
	\; \vxi_i \sim \gN[\vzero, \alpha_i\mI],
    \; a_i > 0. 
\end{align*}

This is closely related to Kalman optimization \cite{6963359}, which instead considers the more general optimization problem of $\nabla\vf\,{=}\,0$, but uses Gaussian inference in the same fashion. 
The posterior update evokes the the Levenberg-Marquadt algorithm \cite{Flet87}, a damped version of Gauss-Newton optimization, due to the regularization $\textstyle\frac{1}{\alpha}\mSigma_{i}\inv$ from the prior
\begin{align*}
	\vmu_{i+1} = 
	\vmu_{\vx\mid\vy}
	&= \vmu_{i} + \mSigma_{\vx\vy}\mSigma_{\vy\vy}\inv(\vy^* - \hat{\vf}(\vmu_{i})),\\
	&\approx \vmu_{i} + (\textstyle\frac{1}{\alpha}\mSigma_{i}\inv + \mJ_{i}\tran\mJ_{i})\inv\mJ_{i}\tran(\vy^* - \hat{\vf}(\vmu_{i})).
\end{align*}

Here, the crucial differences between iterative inference and EM can be made apparent.
EM estimates $\mJ_i$ during the forward pass, while iterative inference linearizes about $p(\vx|\vy)$, requiring the additional hyperparameter $\theta$ and constructing a locally linear model for the surrogate model $\hat{\vf}$.
\begin{align*}
	\text{(It.) } \hat{\vf}(\vx) &= \vf(\hat{\vx}_{i}) + \mJ_i(\vx{-}\hat{\vx}_{i{-}1}),
	&\hat{\vx}_{i} &= \theta \hat{\vx}_{i}{+}(1{-}\theta) \vmu_{\vx|\vy}
	\\
	\text{(EM) } \hat{\vf}(\vx) &= \vf(\vx),
	& \vx_{i} &\sim p(\vx|\vy).
\end{align*}

\begin{figure}[!tb]
	\input{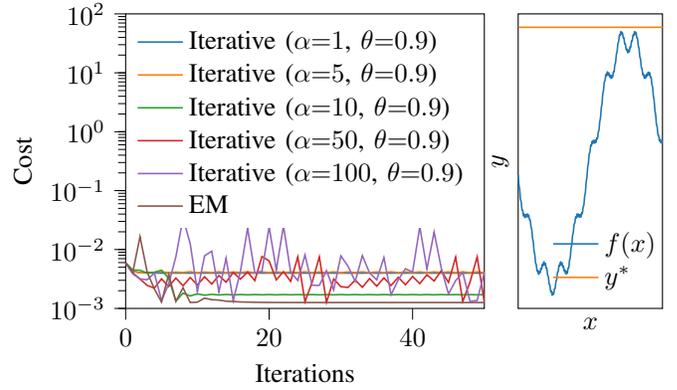}
	
	\vspace{-0.75em}
	\caption{For a 1D non-covex, nonlinear inverse problem, the adaptive regularization provided by the EM update rule can be shown to provide superior stability and convergence compared to fixed hyperparameters for iterative inference.}
	\label{fig:kalman_opt}
\end{figure}

Moreover, the need to iteratively tune $\alpha$ matches the similar heuristics used in Levenberg-Marquadt, and the behavior of \Eqref{eq:alpha_update} also acts to reduce regularization as the cost reduces.  
Using a fixed $\alpha$ likely results in sub-optimal optimization without tuning.
Posterior linearization also poses a practical problem for trajectory optimization, as the subsequent solution is not guaranteed to be close to the previous candidate. 
Therefore, a posterior linearization approach may likely produce an inaccurate state-action trajectory estimate compared to using the true function during filtering.  
To illustrate the importance of these differences empirically, we show the performance on a nonlinear, non-convex inverse problem shown in Figure \ref{fig:kalman_opt}.
Comparing the equivalent approaches used by \itwoc to AICO and PPI, we compare a fixed alpha with smoothing parameter $\theta$ to the EM approach (\Eqref{eq:alpha_update}).
Sweeping through inverse temperatures, the solutions transition from sub-optimal to unstable.
The EM approach automatically calculates adequate value for $\alpha$ to achieve superior convergence.

\subsection{Note on Convergence}
As a damped Gauss-Newton method, Gaussian \itwoc can be analyzed for local convergence under some conditions, but in general does not have guarantees for convergence \cite{bell1994iterated}.
A convergence analysis relates closely to that of Gauss-Newton methods \cite{Flet87}, and more closely to the posterior linearization filter ~\cite{7515187}, which performs accurate Bayesian smoothing of Gaussian densities by iteratively re-linearizing about the posterior estimate.
However, compared to state estimation, trajectory optimization is often initialized further from optimal trajectory, as the cost setpoints $\vz_t^*$ are often less informative of the optimal trajectory than sensor measurements, unless $\vz_t^*$ describes a tracking problem.
Therefore, the convergence requirement that the initial solution is `sufficiently' close to the fixed point is harder to guarantee in this setting, which reinforces the importance of the control priors and $\alpha$ to regularize against divergence.
The posterior linearization filter analysis also highlights the importance of accurate inference for estimating the local likelihood curvature.

\section{Interpreting Inference-based Control}
\label{sec:inference_as_control}
Having demonstrated that Gaussian smoothing performs optimal control, we now discuss the additional benefits of inference-based control. 
This is illustrated by the likelihood objective (\Eqref{eq:i2cmodel}), which contains additional regularization terms which can be connected to prior methods in control.

\subsubsection{Uncertainty-based Regularization}
\label{sec:dual_control}
An important aspect to \itwoc{} is how uncertainty influences the optimized control law.
Most notable from \Eqref{eq:i2c_gaussian_K} is how the gain is attenuated by $\mSigma_{\vx_t}$, indicating that control action reduces with state uncertainty.
This finding parallels the `turn-off phenomenon' observed in dual control \cite{aoki1967optimization,bar1981stochastic} and Bayesian reinforcement learning \cite{DBLP:journals/jmlr/KlenskeH16}, where actions are similarly attenuated under model, and consequently state, uncertainty.
For dual control, this regularization is detrimental due to the active learning component, however this behavior can be useful in settings such as model-based reinforcement learning \cite{deisenroth2013survey}, where regions of predictive uncertainty can indicate modeling error, and uncertainty-based regularization prevents these errors being exploited during optimization \cite{schneider1997exploiting}.

\subsubsection{Risk-Seeking Optimization}
\label{sec:i2c_risk}
Exponentiating the optimal control objective for risk sensitivity (\Eqref{eq:risk})  has a clear connection to control-as-inference, due to the likelihood definition in \Eqref{eq:cost_exponential}.
As previous work has noted, this relates the risk coefficient $\sigma$ to the inverse temperature $\alpha$\cite{rawlik2013probabilistic,levine2013variational,o2018variational}.
An immediate consequence of this is that $\sigma{\,>\,}0$ for control-as-inference, so it is inherently `risk seeking'.
Characterizing risk-seeking behavior as control attenuation, we can interpret this as a manifestation of the turn-off phenomena, as increasing $\sigma$ acts to increase the state uncertainty.
Interestingly, this presents risk-averse control as a counter to the turn-off phenomena, artificially reversing the regularization through the sign reversal of $\sigma$. 
The artificial nature of this mechanism is reflected in the numerical brittleness of risk-averse control, as it requires    
$\sigEta{t}{\,+\,}\textstyle\frac{1}{\sigma}\mV_t\inv$
to be positive definite when $\sigma{\,<\,}0$, which is difficult to ensure.
Beyond attenuating feedback gains, \itwoc uses risk-seeking control in a secondary fashion due to the forward optimization during filtering.
From the M-step update of \itwoc (\Eqref{eq:alpha_update}), we can see that as the average cost decreases, $\alpha$ increases and therefore the effective risk-seeking increases. 
To appreciate why it is beneficial for $\alpha$ to increase during optimization, and not simply remain fixed to a small, more risk neutral value, we must consider the expected log-likelihood objective. 
As $\alpha$ scales the control objective in the overall objective, a small $\alpha$ does not encourage optimization of the cost, while
setting $\alpha$ high leads to overly aggressive forward optimization that neglects the dynamics likelihood, producing unfeasible trajectories. 
This `optimistic' greediness, introduced in Section~\ref{sec:em_inference}, can prefer trajectories that violate the dynamics constraint, as $p(\vx_{t+1}|\vx_t,\vu_t, \vz_{1:t}){\,\neq\,}p(\vx_{t+1}|\vx_t, \vu_t)$~\cite{levine2018reinforcement}.
Conversely, the risk-neutral case turns off this forward optimization, removing the exploration mechanism. 
Therefore, \itwoc leverages risk-seeking optimization to accelerate convergence and the M-step can temper this exploration through adjusting $\alpha$.

\subsubsection{Optimism as Exploration}
\label{sec:exploration_inf}

In this section, we briefly discuss the exploration benefits of inference-based control.
The filtering operation during the forward pass has an exploratory effect, due to the interaction between the state uncertainty and the control cost when the state-action distribution is conditioned on the cost-as-likelihood.
As a result, in the case of \emph{epistemic} model uncertainty, this mechanism will prefer low-cost, uncertain regions of the state space.
This strategy is reminiscent of the upper confidence bound (UCB) for `optimism under uncertainty', an exploration strategy for bandit problems and Bayesian optimization (BO) \cite{bo_bandit}.
When performing BO with a Gaussian process (GP), UCB optimizes for a confidence interval of the uncertain objective, defined through hyperparameter $\beta$.
To compare the exploration strategies of UCB and filtering, 
we consider a simple bandit-like setting of optimizing a point estimate control $u_0$ for one timestep, given a GP dynamics model $f(x,u) {\,=\,}\gN(\mu(x,u),\sigma^2(x,u))$ and known starting state $x_0$ for a $x_1^2$ objective.
The UCB  objective is 
\begin{align*}
	&\min_{u_0} &
	&\E[f(x_0, u_0)^2] - \beta \sqrt{\mathbb{V}[f(x_0, u_0)^2]}.
\end{align*}

\begin{figure}[!tb]
	\vspace{-0.75em}
	\begin{minipage}{\textwidth}
		\input{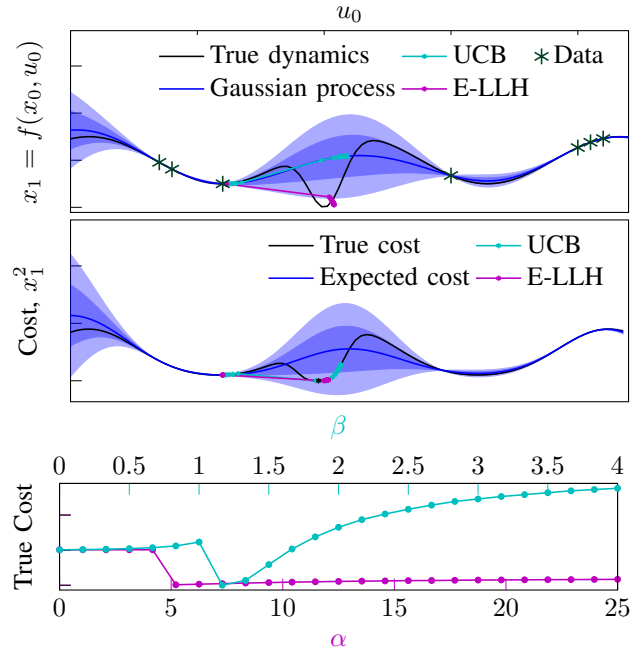}
		\vspace{-0.5em}\\
		\begin{tikzpicture}

\pgfplotsset{ every non boxed x axis/.append style={x axis line style=-} }

\definecolor{color0}{rgb}{0.75,0,0.75}
\definecolor{color1}{rgb}{0,0.75,0.75}

\begin{axis}[
height=3cm,
width=9cm,
tick align=inside,
tick pos=left,
x grid style={white!69.0196078431373!black},
xlabel={\(\displaystyle \alpha\)},
xmin=0, xmax=25,
xtick style={color=color0},
xlabel style={color0},
y grid style={white!69.0196078431373!black},
ylabel={True Cost},
yticklabels={,,},
ymin=-0.137087375570062, ymax=2.90051658356631,
ytick style={color=color0},
]
\addplot [semithick, color0, mark=*, mark size=1, mark options={solid}]
table {%
0.001 1.00886151989694
1.042625 1.00937837588543
2.08425 1.00964232265491
3.125875 1.01018125847157
4.1675 1.01045624944801
5.209125 0.0128916829082347
6.25075 0.0295279140795015
7.292375 0.0455874527103482
8.334 0.0616767428294917
9.375625 0.0751411705833737
10.41725 0.0882617355758067
11.458875 0.0978704529951748
12.5005 0.108202314615081
13.542125 0.116443071575925
14.58375 0.122177816402841
15.625375 0.131150134355664
16.667 0.134241382351573
17.708625 0.140577205162243
18.75025 0.147120345225558
19.791875 0.150470732968031
20.8335 0.15387423470014
21.875125 0.157331274566323
22.91675 0.160842274642519
23.958375 0.164407654874305
25 0.168027833015322
};
\end{axis}

\begin{axis}[
height=3cm,
width=9cm,
axis x line=top,
tick align=inside,
x grid style={white!69.0196078431373!black},
xlabel={\(\displaystyle \beta\)},
xmin=0, xmax=4,
yticklabels={,,},
xtick pos=right,
xtick style={color=color1},
xlabel style={color1},
y grid style={white!69.0196078431373!black},
ymin=-0.137087375570062, ymax=2.90051658356631,
ytick pos=left,
ytick style={color=black},
axis line style={-},
]
\addplot [semithick, color1, mark=*, mark size=1, mark options={solid}]
table {%
0 1.00860860882545
0.166666666666667 1.01569550106053
0.333333333333333 1.02621839189368
0.5 1.04335153467687
0.666666666666667 1.07058803356113
0.833333333333333 1.11951972448555
1 1.22931796999692
1.16666666666667 0.000985531663409662
1.33333333333333 0.137383594770375
1.5 0.58760244477858
1.66666666666667 1.03197334137447
1.83333333333333 1.3771298081449
2 1.65090020595786
2.16666666666667 1.8675516759885
2.33333333333333 2.02939388800004
2.5 2.16278596430287
2.66666666666667 2.28215392806428
2.83333333333333 2.38630167071685
3 2.4566145826412
3.16666666666667 2.52759273825983
3.33333333333333 2.58124210000226
3.5 2.63523136080181
3.66666666666667 2.68954553539941
3.83333333333333 2.72592802434372
4 2.76244367633284
};
\end{axis}

\end{tikzpicture}
	\end{minipage}
	\caption{A simple comparison between exploration using UCB and the expected log-likelihood (E-LLH) for a 1D Gaussian process dynamics model.
		UCB seeks the most uncertain region in the objective, while the inference-strategy seeks $x_1{\,=\,}0$ due to the cost-as-likelihood transformation and filtering step, such that the posterior $x_1$ drifts from the mean prediction.
		As a result, in this scenario, the inference approach selects an effective $u_0$ under uncertainty across a range of $\alpha$ values, while UCB tends to over-explore as $\beta$ increases.
		Uncertainty intervals illustrate one and two standard deviations.
	}
	\label{fig:ucb_vs_ellh}
\end{figure}
The inference approach involves filtering, computing the posterior $p(x_1)$,and optimizing the expected log-likelihood w.r.t. $u_0$ results in the following objective in $u_0$, 
\begin{align*}
	&\min_{u_0} &
	&\alpha^2(\mu_{x_1}^2 {+} \sigma_{x_1}^2)
	{+}
	\textstyle\frac{1}{\sigma^2(x_0,u_0)}(\sigma_{x_1}^2 {+} (\mu_{x_1} {-}\mu(x_0,u_0))^2),\\
\end{align*}

\vspace{-4em}

\begin{align*}
	\gN(\mu_{x_1}, \sigma^2_{x_1})
	&=
	\gN\left(\frac{\mu(x_0,u_0)}{\alpha\sigma^2(x_0,u_0) + 1}, \frac{\sigma^2(x_0,u_0)}{\alpha\sigma^2(x_0,u_0) + 1}\right).
\end{align*}
If $\alpha\sigma^2(x_0,u_0) \approx 0$, this objective minimizes the expected cost.
As $\alpha\sigma^2(x_0,u_0)$ becomes large, optimizing for $u_0$ is no longer possible, as the state posterior will already be close to the optimal state by greedily exploiting uncertainty. 
Therefore, this objective considers the explorations-exploitation trade-off by balancing seeking sources of uncertainty to achieve greedy posterior updates against the resulting expected cost of the trajectory.
Figure \ref{fig:ucb_vs_ellh} shows how UCB and inference-based optimization differ in exploration strategy.
While both methods optimize for the expected cost when exploration is not encouraged, UCB seeks high variance regions of the objective as $\beta$ is increased.
Conversely, in the inference setting $u_0$ is chosen so the state posterior achieves $x_1 {\,=\,} 0$ due to the informative structure of the cost likelihood.

\subsubsection{Maximum Entropy Regularization}
\label{sec:max_ent_inf}
The Gaussian assumption naturally incorporates its log-normalization terms into the log-likelihood, which for the Gaussian distribution is also its negative entropy.
Augmenting the control objective with entropies evokes the maximum entropy (ME) principle, which has been applied to control for robustness, in particular for the controller in inverse optimal control \cite{ziebart2010modeling}.  
The latent Gaussian prior on the controls adds ME regularization to \itwoc{}, which can be derived by examining the posterior policy.
In explicit ME iLQG, the policy covariance takes the form $\mSigma_t{=}(\mC_{\vu_t}{+}\mF_{\vu_t}\tran\mV_{t+1} \mF_{\vu_t})\inv$~\cite{levine2014motor}.
We can see from $Q$ update in \Eqref{eq:Q_update} that this is the inverse of $\mQ_{\vu_t}$, and the expression for $\mSigma_{\vu_t}$ in \itwoc{} (\Eqref{eq:i2c_covariance_control}) corresponds to the Schur complement form of $\mLambda_{\vu_t}$ in \Eqref{eq:precision_joint}.
As the stochastic aspect of the linear Gaussian controller is undesired in practice, the value of this ME regularization could be questioned.
However, ME regularization was observed to influence desirable optimization behavior. 
For robust regions (e.g. where controls are clamped), the control variance was large, whereas at sensitive regimes (e.g. during energy injection in swing-up tasks) this variance was at a minimum.
When updating $\vu$, a lower variance prior reduces the magnitude of the update, therefore for the robust and sensitive regions of the trajectory the control updates are regularized appropriately.

The $\alpha$ update in \Eqref{eq:alpha_update} also has a ME interpretation.
Following van Campenhout et al.\cite{1056374}, we can use the maximum entropy principle to motivate the cost likelihood.
\begin{lemma}\label{th:maxent}
	(Maximum entropy distributions, Section 12.1 \cite{Cover2006})
	Let function $\vh(\vx){\,:\,}\mathbb{R}^{d_x}{\,\rightarrow\,}\mathbb{R}^h$ contain all `useful' information about random variable $\vx$.
	Given an observed empirical average $\hat{\vh}$, and wish to find the density $q(\vx)$ such that $\int q(\vx) \vh(\vx) d\vx{\,=\,}\hat{\vh}$.
	The maximum entropy distribution takes the form $q(\vx){\,=\,}Z^{-1}\exp(\bm{\lambda}\tran\vh(\vx))$.
\end{lemma}
\begin{proposition}\label{th:linearquadratic}
	Following Lemma \ref{th:maxent},
	when the $\vh$ is a linear quadratic potential $h(\vx){\,=\,}\mahalanobis{\vz{\,-\,}\mA\vx}{\mQ}$, where $\vz{\,\in\,}\mathbb{R}^{d_z}$ and $\mQ$ is symmetric positive definite,
	$q(\vx)$ is a multivariate Normal distribution.
	When $\mA$ is invertible, the expectation constraint can be satisfied when $\lambda{\,=\,}d_z/\hat{h}$.
\end{proposition}
This result matches \Eqref{eq:alpha_update} for a single timestep with $\alpha{\,=\,\lambda}$, $h$ as the control cost and $\hat{h}$ obtained from the latent state-action trajectory. 
For the proof, refer to the Appendix.
This perspective reflects how the cost is used to `summarize' the state-action distribution.
Moreover, $\alpha$ is introduced to satisfy a constraint, like in KL-regularized control \cite{deisenroth2013survey,Levine_gps}.

\section{Extensions \& Applications}
\label{sec:extensions}
This section presents several improvements and extensions to \itwoc for the stochastic control setting.

\begin{figure}[!tb]
	\vspace{-2em}
	\begin{minipage}[c]{0.4\columnwidth}
		\begin{tikzpicture}

\pgfplotsset{ticks=none}

\definecolor{color0}{rgb}{0.75,0,0.75}
\definecolor{color1}{rgb}{0,0.75,0.75}

\begin{axis}[
height=5cm,
width=5cm,
legend cell align={left},
legend pos=north west,
legend style={draw=none, fill=none},
tick align=outside,
tick pos=left,
title={$p(u,x)$},
x grid style={white!69.01960784313725!black},
xlabel={$x$},
xmin=-4, xmax=4,
xtick style={color=black},
y grid style={white!69.01960784313725!black},
ylabel={$u$},
ymin=-4, ymax=4,
ytick style={color=black}
]
\draw[draw=blue,fill opacity=0,rotate around={45:(axis cs:0,0)}] (axis cs:0,0) ellipse (1.37840487520902 and 0.316227766016838);
\draw[draw=blue,fill opacity=0,rotate around={45:(axis cs:0,0)}] (axis cs:0,0) ellipse (2.75680975041804 and 0.632455532033675);
\draw[draw=blue,fill opacity=0,rotate around={45:(axis cs:0,0)}] (axis cs:0,0) ellipse (4.13521462562707 and 0.948683298050513);
\addplot [semithick, color0, solid]
table {%
-4 -4
-3.91919191919192 -3.91919191919192
-3.83838383838384 -3.83838383838384
-3.75757575757576 -3.75757575757576
-3.67676767676768 -3.67676767676768
-3.5959595959596 -3.5959595959596
-3.51515151515152 -3.51515151515152
-3.43434343434343 -3.43434343434343
-3.35353535353535 -3.35353535353535
-3.27272727272727 -3.27272727272727
-3.19191919191919 -3.19191919191919
-3.11111111111111 -3.11111111111111
-3.03030303030303 -3.03030303030303
-2.94949494949495 -2.94949494949495
-2.86868686868687 -2.86868686868687
-2.78787878787879 -2.78787878787879
-2.70707070707071 -2.70707070707071
-2.62626262626263 -2.62626262626263
-2.54545454545455 -2.54545454545455
-2.46464646464646 -2.46464646464646
-2.38383838383838 -2.38383838383838
-2.3030303030303 -2.3030303030303
-2.22222222222222 -2.22222222222222
-2.14141414141414 -2.14141414141414
-2.06060606060606 -2.06060606060606
-1.97979797979798 -1.97979797979798
-1.8989898989899 -1.8989898989899
-1.81818181818182 -1.81818181818182
-1.73737373737374 -1.73737373737374
-1.65656565656566 -1.65656565656566
-1.57575757575758 -1.57575757575758
-1.49494949494949 -1.49494949494949
-1.41414141414141 -1.41414141414141
-1.33333333333333 -1.33333333333333
-1.25252525252525 -1.25252525252525
-1.17171717171717 -1.17171717171717
-1.09090909090909 -1.09090909090909
-1.01010101010101 -1.01010101010101
-0.929292929292929 -0.929292929292929
-0.848484848484848 -0.848484848484848
-0.767676767676767 -0.767676767676767
-0.686868686868686 -0.686868686868686
-0.606060606060606 -0.606060606060606
-0.525252525252525 -0.525252525252525
-0.444444444444444 -0.444444444444444
-0.363636363636363 -0.363636363636363
-0.282828282828282 -0.282828282828282
-0.202020202020202 -0.202020202020202
-0.121212121212121 -0.121212121212121
-0.0404040404040402 -0.0404040404040402
0.0404040404040407 0.0404040404040407
0.121212121212122 0.121212121212122
0.202020202020202 0.202020202020202
0.282828282828283 0.282828282828283
0.363636363636364 0.363636363636364
0.444444444444445 0.444444444444445
0.525252525252526 0.525252525252526
0.606060606060606 0.606060606060606
0.686868686868687 0.686868686868687
0.767676767676768 0.767676767676768
0.848484848484849 0.848484848484849
0.92929292929293 0.92929292929293
1.01010101010101 1.01010101010101
1.09090909090909 1.09090909090909
1.17171717171717 1.17171717171717
1.25252525252525 1.25252525252525
1.33333333333333 1.33333333333333
1.41414141414141 1.41414141414141
1.4949494949495 1.4949494949495
1.57575757575758 1.57575757575758
1.65656565656566 1.65656565656566
1.73737373737374 1.73737373737374
1.81818181818182 1.81818181818182
1.8989898989899 1.8989898989899
1.97979797979798 1.97979797979798
2.06060606060606 2.06060606060606
2.14141414141414 2.14141414141414
2.22222222222222 2.22222222222222
2.3030303030303 2.3030303030303
2.38383838383838 2.38383838383838
2.46464646464647 2.46464646464647
2.54545454545455 2.54545454545455
2.62626262626263 2.62626262626263
2.70707070707071 2.70707070707071
2.78787878787879 2.78787878787879
2.86868686868687 2.86868686868687
2.94949494949495 2.94949494949495
3.03030303030303 3.03030303030303
3.11111111111111 3.11111111111111
3.19191919191919 3.19191919191919
3.27272727272727 3.27272727272727
3.35353535353535 3.35353535353535
3.43434343434344 3.43434343434344
3.51515151515152 3.51515151515152
3.5959595959596 3.5959595959596
3.67676767676768 3.67676767676768
3.75757575757576 3.75757575757576
3.83838383838384 3.83838383838384
3.91919191919192 3.91919191919192
4 4
};
\addlegendentry{Linear}
\addplot [semithick, color1, solid]
table {%
-4 -4.50140698877036e-07
-3.91919191919192 -8.36382485483423e-07
-3.83838383838384 -1.53322210163533e-06
-3.75757575757576 -2.77294238195166e-06
-3.67676767676768 -4.94770595905558e-06
-3.5959595959596 -8.70934008609851e-06
-3.51515151515152 -1.51243044053764e-05
-3.43434343434343 -2.59098058272497e-05
-3.35353535353535 -4.37865332181881e-05
-3.27272727272727 -7.29949617737825e-05
-3.19191919191919 -0.000120035171260331
-3.11111111111111 -0.000194703549948027
-3.03030303030303 -0.000311511646827263
-2.94949494949495 -0.000491579685202087
-2.86868686868687 -0.000765095448568764
-2.78787878787879 -0.00117441265984431
-2.70707070707071 -0.00177782480306558
-2.62626262626263 -0.00265398339091973
-2.54545454545455 -0.0039068275383199
-2.46464646464646 -0.00567075042146195
-2.38383838383838 -0.00811554853285565
-2.3030303030303 -0.0114504895108549
-2.22222222222222 -0.0159266111946942
-2.14141414141414 -0.0218361571007936
-2.06060606060606 -0.0295079018171114
-1.97979797979798 -0.0392970733747306
-1.8989898989899 -0.0515686925700456
-1.81818181818182 -0.0666734722067343
-1.73737373737374 -0.0849159889418365
-1.65656565656566 -0.106515666983763
-1.57575757575758 -0.131562167760769
-1.49494949494949 -0.159967986391643
-1.41414141414141 -0.191422286621117
-1.33333333333333 -0.225351087208088
-1.25252525252525 -0.260889640935424
-1.17171717171717 -0.296873015298732
-1.09090909090909 -0.331850315838963
-1.01010101010101 -0.364126583564483
-0.929292929292929 -0.391834147622911
-0.848484848484848 -0.413032255164161
-0.767676767676767 -0.425830401946563
-0.686868686868686 -0.428527340451309
-0.606060606060606 -0.419754716712526
-0.525252525252525 -0.398612167198035
-0.444444444444444 -0.36477991479923
-0.363636363636363 -0.318595719596027
-0.282828282828282 -0.261085542183643
-0.202020202020202 -0.193941300857078
-0.121212121212121 -0.119444245818205
-0.0404040404040402 -0.0403381351636475
0.0404040404040407 0.040338135163648
0.121212121212122 0.119444245818205
0.202020202020202 0.193941300857079
0.282828282828283 0.261085542183643
0.363636363636364 0.318595719596028
0.444444444444445 0.36477991479923
0.525252525252526 0.398612167198035
0.606060606060606 0.419754716712526
0.686868686868687 0.428527340451309
0.767676767676768 0.425830401946563
0.848484848484849 0.413032255164161
0.92929292929293 0.391834147622911
1.01010101010101 0.364126583564483
1.09090909090909 0.331850315838963
1.17171717171717 0.296873015298732
1.25252525252525 0.260889640935424
1.33333333333333 0.225351087208088
1.41414141414141 0.191422286621117
1.4949494949495 0.159967986391643
1.57575757575758 0.131562167760769
1.65656565656566 0.106515666983762
1.73737373737374 0.0849159889418363
1.81818181818182 0.0666734722067342
1.8989898989899 0.0515686925700455
1.97979797979798 0.0392970733747305
2.06060606060606 0.0295079018171113
2.14141414141414 0.0218361571007935
2.22222222222222 0.0159266111946942
2.3030303030303 0.0114504895108549
2.38383838383838 0.00811554853285563
2.46464646464647 0.00567075042146193
2.54545454545455 0.00390682753831989
2.62626262626263 0.00265398339091972
2.70707070707071 0.00177782480306557
2.78787878787879 0.00117441265984431
2.86868686868687 0.00076509544856876
2.94949494949495 0.000491579685202086
3.03030303030303 0.000311511646827261
3.11111111111111 0.000194703549948026
3.19191919191919 0.00012003517126033
3.27272727272727 7.29949617737822e-05
3.35353535353535 4.3786533218188e-05
3.43434343434344 2.59098058272496e-05
3.51515151515152 1.51243044053764e-05
3.5959595959596 8.70934008609845e-06
3.67676767676768 4.94770595905556e-06
3.75757575757576 2.77294238195166e-06
3.83838383838384 1.53322210163532e-06
3.91919191919192 8.3638248548342e-07
4 4.50140698877036e-07
};
\addlegendentry{Expert}
\end{axis}

\end{tikzpicture}
	\end{minipage}\hfill
	\begin{minipage}[c]{0.5\columnwidth}
		\vspace{-.5em}
		\caption{The conditional Gaussian distribution as a linear control law.
			The standard linear control applies far outside the expected distribution, where it was not designed for. The `expert' controller reverts to the prior when far from the mean, which prevents erroneous feedback control.}
		\label{fig:expert}
	\end{minipage}
	\vspace{-2em}
\end{figure}
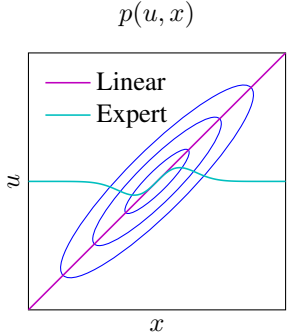

\subsection{Expert Linear Gaussian Controllers}
\label{sec:expert}

A weakness of the Gaussian assumption made in Section \ref{sec:gaussian} is the local nature of the estimate.
Gaussian inference is equivalent to regularized local linearizations about the mean state-action trajectory.
The local nature of this assumption introduces a brittleness to the control that limits its application in practical settings without additional modifications like replanning, a limitation it shares with linearization-based control algorithms.
Using inference, we can leverage the state belief to reduce this brittleness.
In statistical machine learning, an \emph{expert} is a model whose appropriateness applies to a specific portion of a state space \cite{gormley2019mixture}.
We can incorporate this idea into the controller by switching between open- and closed-loop control using the predicted state distribution, (Figure \ref{fig:expert}),  
\begin{align*}
	\vpi_t(\vu|\vx) &= p_{\text{CL}}\,p(\vu_t\mid\vx_t) + (1{-}{p_\text{CL}})p(\vu_t),
	\,\;p_{\text{CL}}{\,=\,}\text{Pr}(\vx_t{=}\vx).
\end{align*}
For continuous random variables, $\text{Pr}(\vx_t{=}\vx)$ is an inconvenient quantity to compute.
We take inspiration from outlier detection and define a suitable confidence interval.
For a multivariate Normal distribution, our $\text{Pr}(\vx_t{=}\vx)$
can be computed using the Mahalanobis distance $d(\vx){\,=\,}\lVert\vx-\mum{\vx}{}{t}\rVert_{\mSigma_{\vx_t}^{-1}}^2$, which has a chi-square distribution.
Using the cumulative density function $F_{\mathcal{X}^2_k}$,
$\text{Pr}(\vx_t{=}\vx){\,=\,}1{\,-\,}F_{\mathcal{X}^2_k}(d(\vx))$ \cite{GVK330798693}.
For $k{\,=\,}2$, $\text{Pr}(\vx_t{=}\vx){\,=\,}\exp(-\textstyle\frac{1}{2}d(\vx))$, which works well and is convenient to compute as the unnormalized density of $\vx_t$.

While this expert controller is valuable during optimization, it can also be used in execution, as it provides a degree of `safety' by turning off feedback.
However, there are also situations where this feedback relaxation impedes performance, such as unstable systems where high feedback is critical.
\begin{figure}[!tb]
	\vspace{-2em}
	\begin{minipage}[t][0.8cm][t]{\columnwidth}
		\hspace{1.4cm}
		\begin{tikzpicture}

\begin{axis}[
hide axis,
width=8.5cm,
xmin=10, xmax=50,
ymin=0, ymax=1.0,
legend cell align={center},
legend columns=4,
legend style={/tikz/every even column/.append style={column sep=0.3cm}, draw=none},
]
]

\addlegendimage{semithick, black, mark=x, mark size=3, mark options={solid}, only marks}
\addlegendentry{$\mathbf{x}_0$};
\addlegendimage{semithick, red, mark=x, mark size=3, mark options={solid}, only marks}
\addlegendentry{$\mathbf{x}_g$};
\addlegendimage{semithick, green!50.0!black, mark=*, mark size=1, mark options={solid}}
\addlegendentry{Closed-loop};
\addlegendimage{semithick, color0, opacity=0.5, dashed, mark=*, mark size=1}
\addlegendentry{Rollouts};
\end{axis}

\end{tikzpicture}
	\end{minipage}
	\begin{minipage}[t][5.8cm][t]{\columnwidth}
		\centering
		\input{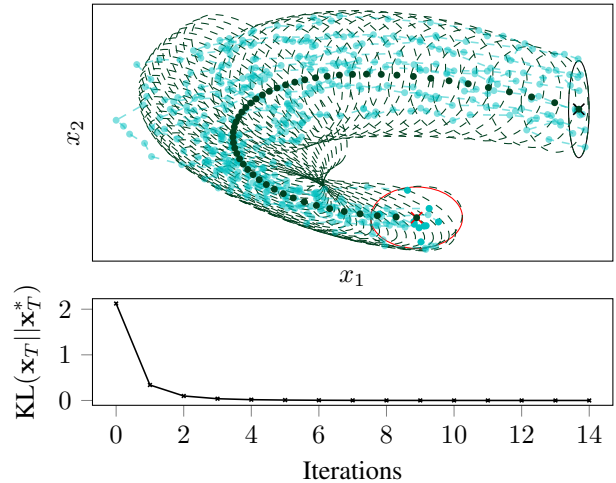}
	\end{minipage}
	\caption{\itwoc{} for exact minimum-energy linear Gaussian covariance control on an unstable system, with a fixed small $\alpha$ and $\sigEta{t}{=}\text{diag}(0.1, 0.1)$.
		The KL divergence is between the terminal goal and closed-loop distributions.}
	\label{fig:covariance}
\end{figure}

\begin{figure*}[!tb]
	\vspace{-2em}
	\includegraphics{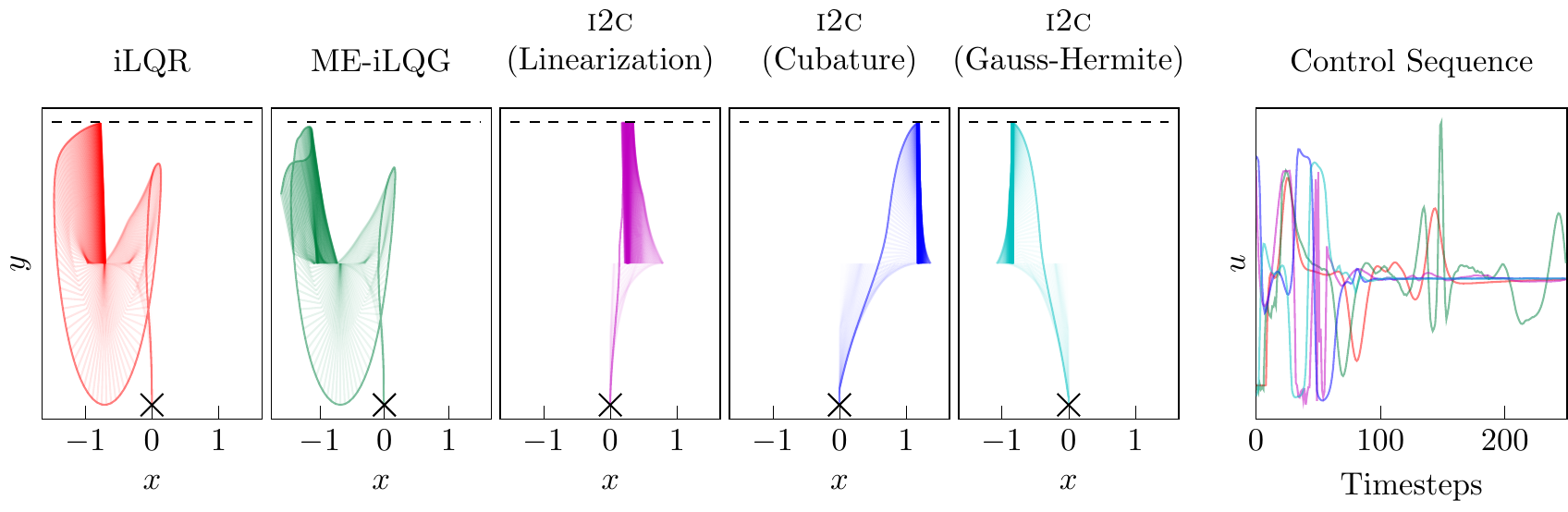}
	\vspace{-1em}
	\caption{Optimized trajectories of the double cartpole tip position, starting from $\times$. Note that the \itwoc variants return similar results due to shared hyperparameters.
		While the quadrature methods have cleaner trajectories due to the accuracy of their inference, that are remarkably similar given Gauss-Hermite requires much greater computation.
	}
	\label{fig:dcp_traj}
	\vspace{-1.5em}
\end{figure*}

\subsection{Covariance Control as Inference}
\label{sec:cov_control_inf}
\begin{figure}[!b]
	\vspace{0.3em}
	\begin{minipage}[t][2cm][t]{\columnwidth}
		\centering
		\input{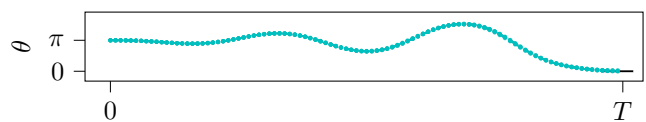}
	\end{minipage}
	\vspace{-1.5em}
	\caption{Nonlinear minimum-energy covariance control on the pendulum swing-up task, using \itwoc with approximate inference. Plot depicts the inferred trajectory~\ref{plan} for target distribution \ref{target}, with simulated rollouts \ref{rollout}.}
	\label{fig:cc}
\end{figure}

In the standard \itwoc{} likelihood objective, the terminal cost is defined with a dedicated observation model $p(\vz_T\mid\vx_T)$.
However, during inference we are also free to directly set the terminal latent state distribution before smoothing.
In other timeseries inference settings such as state estimation, the terminal state posterior is set to the prior as there is no additional data to draw upon.
However, for control we can avoid cost function design and the cost-to-likelihood translation by setting the terminal distribution directly.
A downside of this approach is the requirement to stipulate the desired state directly rather than through a useful transformation that the cost usually provides.
This approach is equivalent to covariance control (Section \ref{sec:cov_control}) as iterations of forward and backward Riccati equations are performed until the boundary condition are satisfied.
In this setting, $\mSigma_{\vxi_T}$ now acts as the Lagrange multiplier.
Examining the expected log-likelihood term for the terminal state for the direct state optimization case $\vz_T{\,=\,}\vx_T{\,+\,}\vxi_T$ \cite{i2cacc}, we see
\begin{align*}
	\E[\log p(\vz_t|\vx_T)|\vx_T]
	&\propto\lVert\mum{\vz}{}{T}{-}\mum{\vx}{}{T}\rVert_{\sigXi}^2{+} \text{tr}\{\mSigma_{\vxi_T}\inv\sigm{\vx}{~}{T}\}+\dots.
\end{align*}
which corresponds to the LQG correspondence proved in Equation 41 of Goldshtein et al. \cite{8264189}, where $\mSigma_{\vxi_T}\inv$ is the terminal cost / Lagrange multiplier matrix.
However, rather than compute this term, using our probabilistic framework we can set the terminal distribution directly via the posterior of $\vx_T$.
In this case, the inference iteratively seeks to satisfy the boundary conditions on $\vx_1$ and $\vx_T$.
Figure \ref{fig:covariance} demonstrates covariance control on a linear Gaussian system, where inference is exact.
Note that \itwoc{} uses stochastic linear Gaussian controllers, whereas the previous literature solves the task using deterministic linear control.
This variation of \itwoc{} naturally translates to nonlinear systems (Figure \ref{fig:cc}), avoiding the complexity of the additional forward sampling required for linearization-based covariance control \cite{yi2019nonlinear}.
However, the terminal boundary constraint still requires a means of being applied in an gradual manner, due to the iterative aspect of the nonlinear optimization.
The terminal state distribution can be shifted from initially near the prior to the desired distribution by `annealing' \cite{annealing} the prior $p(\overrightarrow{\vx}_T)$ each iteration, i.e. $p(\vx_T){\,=\,} p(\vx_T^*)\,p(\overrightarrow{\vx}_T)^{\beta}\text{ where }\beta{\,\rightarrow\,}0$ following a linear schedule over iterations.
\subsection{Integrating State Estimation and Control}
\label{sec:state_estimation}
As \itwoc uses recursive Bayesian estimation, it can be integrated seamlessly with state estimation algorithms, which correspond to the partially-observed optimal control problem.
In the standard LQG problem, the separation principle applies and the combination of a Kalman filter and LQR controller is optimal \cite{stengel1986stochastic}. 
In the nonlinear setting, this convenient separation is no longer valid, however it is still applied in practice \cite{todorov2005generalized}.
For \itwoc, the probabilistic graphical model is now extended to include a measurement model, i.e. $\vy_t\!=\!\vg_t(\vx)\!+\!\vzeta_t$, $\vy\!\in\!\mathbb{R}^{d_y}$, {$\vzeta_t{\,\sim\,}\gN(\vzero, \mSigma_{\vzeta})$}.
Under this time-varying graphical model, the past and present state distribution is obtained using state estimation, so the planned controls must adapt to the updated state distribution at each timestep.

This replanning procedure naturally evokes model predictive control, which is also motivated by adapting optimal control to an evolving state distribution \cite{MORARI1999667}.
There is also a connection to dual control and the notion of closed-loop control vs. feedback control of Tse and Bar Shalom \cite{1100635}, as early work on replanning was motivated by reducing the adverse consequences of the turn-off phenomena (discussed in Section \ref{sec:dual_control}) \cite{1099243,1100635}. 
\begin{table*}[tb]
\vspace{-1em}
\centering
\resizebox{\textwidth}{!}{%
\begin{tabular}{llllllllll}
	\bfseries
	Environment & \multicolumn{9}{c}{\bfseries 10th, 90th Cost Percentiles ($\times 10^3$)}\\
	\cmidrule(lr){1-1} \cmidrule(lr){2-10}
                & \itwoc{} (S, E)& \itwoc{} (CE, E)& \itwoc{} (S, FF) & \itwoc{} (S, FB) & \itwoc{} (CE, FF) & \itwoc{} (CE, FB) & SQP (CE, FF) & iLQR (CE, FB) & ME-iLQG (S, FB)\\
\cmidrule(lr){2-10}
Pendulum, $\vtau{\,\in\,}\sR^3$ & 13.46, 21.53 & \textbf{12.81, 17.11} & 17.72, 21.94 & 19.23, 21.43 & 13.97, 26.77 & 19.49, 22.31 & 18.10, 26.31 & 23.33, 26.46 & 19.45, 20.91\\
Cartpole, $\vtau{\,\in\,}\sR^5$ & 85.06, 87.43 & \textbf{81.83, 83.87} & 89.53, 94.67 & 93.53, 95.71 & 86.93, 89.75 & 121.89, 123.88 & 111.31, 118.57 & 142.23, 145.78 & 120.80, 122.45\\
	\bottomrule
\end{tabular}
}
\caption{
	The evaluation of SOC algorithms on finite-horizon, input-constrained control tasks.
	Variations are characterized by optimizing the stochastic (S) or certainty equivalent (CE) setting and using open-loop (FF), closed-loop (FB) or expert (E) controllers during optimization.
	These features identify similarities in performance.
	Percentiles were computed from 100 rollouts.
	SQP combines an open-loop trajectory optimized with sequential quadratic programming with local LQR feedback.
}
\label{tab:results}
\vspace{-1.5em}
\end{table*}

\section{Simulated Experiments}
\label{sec:result}

This section provides benchmarks of the \itwoc algorithm described in Section \ref{sec:i2c} as a trajectory optimization solver, as well as empirical investigations into the extensions discussed in Section \ref{sec:extensions}: Expert-controllers and partially-observed MPC.

\subsection{Approximate Inference for Control}
\label{sec:approx_inf_exp}

In Section \ref{sec:nonlinear}, several approximate Gaussian message passing methods were discussed in order to apply Gaussian \itwoc to nonlinear systems. 
For this experiment, we evaluate each method on a deterministic double cartpole swing-up task.
As a double cartpole is a chaotic nonlinear system where $\vtau{\,\in\,}\sR^{7}$, the degree of nonlinearity should challenge the accuracy of the approximate inference.
As a baseline, we consider iLQR and maximum-entropy iLQG (denoted ME-iLQG) \cite{Levine_gps}, which use linearization-based approximations with line-search and relative entropy constrained updates respectively. 

\begin{figure}[!b]
	{
		\input{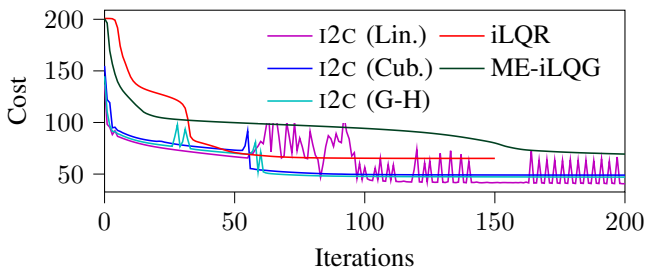}
		\caption{Cost of the mean trajectory during trajectory optimization for double cartpole swing-up. While all \itwoc variants converge on similar solutions, the inaccuracy of linearization-based inference makes optimization more unstable.
		Figure \ref{fig:dcp_cost_alpha} in the Appendix compared \itwoc performance over fixed values of $\alpha$ instead of the EM strategy.}
		\label{fig:dcp_cost}
	}
\end{figure}

In order to demonstrate the consequences of approximate inference, we use the same priors and regularization across all \itwoc variants. 
In doing so, we show that \itwoc variations achieve similar high performance, but illustrate in Figure \ref{fig:dcp_cost} how inference quality impacts optimization. 
Another important consideration is computation time. 
As discussed earlier, linearization-based approaches are unwieldy due to the Jacobian computation, especially for environments where Taylor approximations are not straightforward to compute. 
Moreover, iLQR requires line search for regularization which introduces additional computation. 
ME-iLQG uses cheaper, but usually conservative, KL regularization.
The benefit of \itwoc is the adaptive regularization of the Bayes rule, which comes with the caveat of `optimistic exploration'.
Moreover, the use of inference allows for greater flexibility in computation.
Table~\ref{tab:time_tab} reports the average iteration time for double cartpole optimization.
Note that the implementations of iLQR and ME-iLQG have optimized codebases, with pre-compiled Riccati equations and parallelized Jacobian computation. 
Despite this, iLQR is the most expensive, primarily due to the line search, as iterations were observed to be up to $\times 2$ faster than the average. 
ME-iLQG is faster than linearized \itwoc due to the ability to parallelize linearization, however cubature inference is the significantly faster method, as it balances computation and accuracy in a very attractive fashion in the nonlinear Gaussian setting.
Moreover, while G-H is understandably the more expensive quadrature method, it was found to be faster than iLQR. 
Secondly, while G-H is clearly more accurate than cubature (e.g. Figure \ref{fig:inference}), the results in Figures \ref{fig:dcp_traj} and \ref{fig:dcp_cost} show that for double cartpole cubature gave sufficient accuracy.

\subsection{Trajectory Optimization of Stochastic Systems}
To evaluate the performance of the expert controller (Section~\ref{sec:expert}), we consider two stochastic, nonlinear swing-up tasks against various open- and closed-loop baselines.
The solvers also vary between considering the actual stochastic problem or a certainty equivalent approximation.
In Table \ref{tab:results} we compare  both \itwoc and baseline SOC solvers.
For both tasks, open-loop methods resulted in better optima but were also high variance in the cost, while the closed-loop alternatives had much lower variance but sub-optimal performance on the simulated systems due to their over-actuation.
Reassuringly, the results of the \itwoc variant and equivalent baseline solver were generally similar due to the comparable computation.

Incorporating the expert controller into the E-step softens the feedback control during exploration, effectively acting open-loop.
This avoids highly-actuated trajectories forming, avoiding local optima.
Table \ref{tab:results} demonstrates the effectiveness of this addition, where this expert controller matches the open-loop optima but with the closed-loop variance reduction.

\begin{table}[!b]
	\centering
	{%
		\begin{tabular}{llllll}
			& iLQR & ME-iLQG & \itwoc (Lin.) & \itwoc (Cub.) & \itwoc (G-H) \\  \midrule
			Time (s)  &  1.0 & 0.21 & 0.35 & \textbf{0.03}  & 0.69  \\
			\bottomrule     
		\end{tabular}
	}
	\caption{Relative computation time per iteration for double cartpole trajectory optimization, averaged over a maximum of 200 iterations and normalized about iLQR. G-H is degree 4.}
	\label{tab:time_tab}
\end{table}

\subsection{High-Dimensional Model Predictive Control}

For controlling complex, high-dimensional systems, local iterative solvers will struggle numerically when the task is highly non-convex and linearizations are not explicitly provided. 
We demonstrate the scalability of quadrature-based \itwoc on a challenging humanoid stand-up task, using MPC to amortize the difficult trajectory optimization.
As a baseline, we consider sample-based MPC using the cross-entropy method (CEM) \cite{cem}, an effective black-box optimization algorithm adopted for MPC \cite{Boots-RSS-19}.
CEM optimizes a factorized Gaussian distribution over action sequences, optimizing by moment matching the top performing Monte Carlo rollouts.
The true cost is used, rather than the exponentiated form.
As a result, CEM is free from the stucture imposed by Gaussian message passing computation, but consequently suffers from high variance in the trajectories due to the Monte Carlo rollouts. 
Therefore, CEM provides an indication of the performance of the MAP solution using Monte Carlo computation, which in turn allows us to assess the effectiveness of the Gaussian approximate and quadrature inference of \itwoc. 

\begin{figure}[!tb]
	\centering
	\includegraphics[width=7cm]{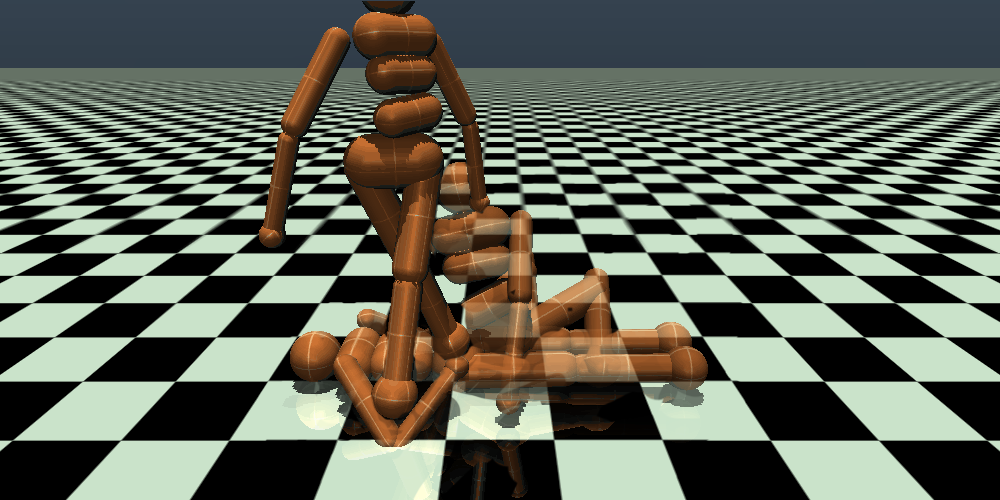}
	\caption{Visualization of the humanoid standing trajectory with \itwoc MPC using cubature quadrature.}
	\label{fig:humanoid}
\end{figure}

The MPC algorithms are evaluated on a humanoid stand-up task using the MuJoCo simulator \cite{mujoco}.
The humanoid has 17 actuators across 5 joints, and its state space (both intrinsic and extrinsic) is represented in $\mathbb{R}^{27}$. The task, depicted in Figure \ref{fig:humanoid}, requires control to stand to a height of $1.25$m from rest in $1.8$s, with a planning horizon of $0.72$s represented as 30 control steps. 
Both methods share a control prior of $\mSigma_\vu{\,=\,} 0.1\mI$, as the actuators are limited to $[-1,1]$, use 1 iteration per timestep and a warm-start of 50 iterations.
For $\itwoc$, $\alpha$ is calibrated w.r.t. the prior and warm-start solution, but is kept fixed during execution as the optimization problem is non-stationary. 
Figure~\ref{fig:humanoid} demonstrates \itwoc MPC can successfully solve the task, and Table \ref{tab:humanoid} shows that quadrature-based \itwoc inference is an effective use of particles, but superior performance can be obtained by CEM when significantly more samples are available.
This suggests that a sequential Monte Carlo for \itwoc inference may be effective for such non-convex tasks.   

\begin{table}[!b]
	\centering
	\resizebox{\columnwidth}{!}
	{%
		\begin{tabular}{llllll}
			& \itwoc & \textsc{CEM} & \textsc{CEM}  & \textsc{CEM} & \textsc{CEM} \\  
			&  $(n{=}88)$ &  $(n{=}50)$ & $(n{=}100)$ & $(n{=}500)$&  $(n{=}1000)$ \\  \midrule
			Cost  & 47.78
			      & 105.28, 108.33 %
			      & 59.61, 62.82 %
			      & 42.55, 55.43 %
			      & 34.09, 51.72 \\ %
			\bottomrule     
		\end{tabular}
	}
	\caption{Cost percentiles (10th, 90th) for the humanoid stand-up task over 25 seeds for MPC solvers with $n$ particles.
	\itwoc is deterministic and requires $2d_{xu}$ points for inference.
	CEM demonstrates superior best-case performance, but only with significantly more rollouts.}
	\label{tab:humanoid}
\end{table}

\subsection{Partially Observed Model Predictive Control}

\begin{figure}[!t]
	\vspace{-.5em}
	\begin{minipage}[t][3cm][t]{\columnwidth}
		\centering
		\input{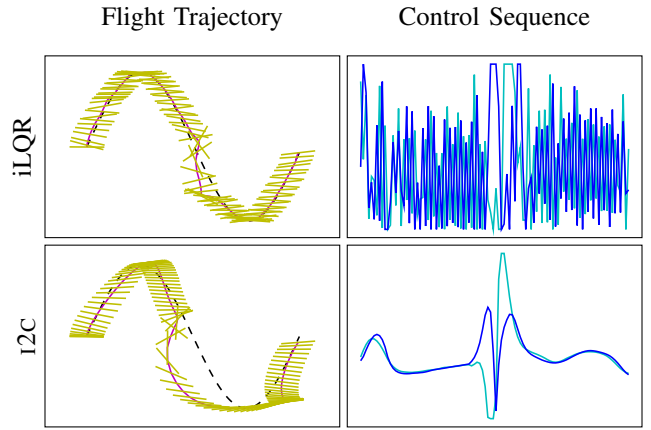}
	\end{minipage}
	\vspace{-1em}
	\caption{Comparing the control of \itwoc and iLQR for an acrobatic quadcopter task, consisting of a smooth trajectory and 360$^\circ$ flip, under high measurement noise ($\mSigma_\zeta^\text{high}$). The trajectory is depicted by the desired and achieved trajectory, along with the measured pose. The controls consist of two thrusts applied to each end of the vehicle.} \label{fig:mpc_compare}
\end{figure}

To investigate \itwoc MPC with state estimation, we compare cubature Gaussian \itwoc with iLQR for a 2D acrobatic quadcopter task ($\vx{\,\in\,}\sR^6$, $\vu{\,\in\,}\sR^2$), using a cubature Kalman filter for state estimation.
Tracking a pre-specified trajectory, we evaluate the performance across increasing measurement uncertainty using the measurement model from Section \ref{sec:state_estimation}.
To simulate interesting state estimation dynamics, the positions and velocities of the left and right side of the copter in the world frame are measured.
Therefore, during the somersault (Figure \ref{fig:mpc_compare}) the state uncertainty fluctuates, as the state is observed in a nonlinear manner.

We evaluate two measurement noise settings: low and high noise ({$\mSigma_\zeta^\text{low}$, $\mSigma_\zeta^\text{high}$}), where high noise masks out the velocities and right side position sensor.
Low noise evaluates the methods for MPC, while the high noise case evaluates the effect of increased state uncertainty on \itwoc.
iLQR is a deterministic, risk-neutral method, while \itwoc is probabilistic and risk-seeking, therefore we aim to understand how \itwoc uses of the state uncertainty from filtering and whether this aids control.
Table \ref{tab:mpc_res} discusses the results.

\section{Conclusion}
\label{sec:conclusion}
We have discussed how input estimation can be used to frame optimal control as an approximate inference problem.
We have focused on the assumption of a Gaussian state-action distribution which, while having limited expressivity, provides the benefits of Gauss-Newton optimization and results in an uncertainty-regularized dynamic programming solver. 
While uncertainty-based regularization comes with (long-established) weaknesses, this approach has been demonstrated to be numerically competitive with popular solvers while also extending to alternative stochastic control methods like covariance control.
Different approximation methods for Gaussian message passing were assessed and evaluated, comparing accuracy, computational cost and downstream control performance.
Future work should consider relaxing the Gaussian assumption through methods such as sequential Monte Carlo \cite{SCHON201139}, enabling multi-modal trajectory optimization.

\begin{table}[!b]
	\centering
	{%
		\begin{tabular}{p{0.25\columnwidth}p{0.25\columnwidth}p{0.25\columnwidth}}
			\textbf{Algorithm} & \multicolumn{2}{c}{\textbf{10th, 90th Cost Percentiles}}\\ \midrule
			& $\mSigma_\zeta^\text{low}$ & $\mSigma_\zeta^\text{high}$ \\ \cmidrule(lr){2-3}
			\itwoc (FF) & 40.59, 41.52             & 113.35, 125.96 \\
			\itwoc (FB) & \textbf{39.98, 41.03}    & 113.41, 126.02 \\
			iLQR   (FF) & 111.73, 1968.39          & 107.02, 1809.36 \\
			iLQR   (FB) & 48.67, 68.35             & \textbf{53.85, 73.33}  \\ \bottomrule	
		\end{tabular}
	}
	\caption{Acrobatic quadrocopter tracking performance under increasing state uncertainty over 50 random seeds.
		In the low noise setting, the \itwoc controller is superior and consistent across control modes.
		In the high noise setting, the increased states uncertainty induces \itwoc to regularize the control, leading to an increased cost that surpasses iLQR, but with reduced spread.
		This task setting illustrates when uncertainty-based regularization can be detrimental to performance. 
	}
	\label{tab:mpc_res}
\end{table}

\newpage

\bibliography{lib}
\bibliographystyle{IEEEtran} 

\newpage

\appendix
\vspace{-0.5em}
Experimental details can be found in the codebase\\ {\scriptsize\url{www.github.com/JoeMWatson/input-inference-for-control}}.

\vspace{-1em}

\subsection{Proof for Proposition 5.1}
\begin{proof}
	When $\vh$  is the linear quadratic potential
	$\mahalanobis{\vz{\,-\,}\mA\vx}{\mQ}$,
	where $\mA$ in invertible and $\mQ$ is symmetric pd,
	$q(\vx)$ is Gaussian with the moments\\
	$
	q(\vx) =
	\gN(\vmu_q, \mSigma_q) = \gN((\bar{\mA}\tran\bar{\mA})\inv\bar{\mA}\tran\mP\vz,\textstyle\frac{1}{\lambda}(\bar{\mA}\tran\bar{\mA})\inv),
	$
	where $\bar{\mA}{=}\mP\mA$, using the Cholesky decomposition of $\mQ{\,=\,}\mP\tran\mP$.
	Following the quadratic form for Gaussian variables, the expected cost under $q$ can be expressed as \cite{quad_forms} \\
	$\E_{\vx\sim q(\cdot)}[h(\vx)]{\,=\,}\tr\{\mQ\mA\mSigma_q\mA\tran\}{\,+\,} (\vz{-}\mA\vmu_q)\tran\mQ(\vz{-}\mA\vmu_q)$.
	The right term is zero as $\mA\vmu_q{\,=\,}\mA(\mP\mA)\inv\mP\vz{\,=\,}\vz$.
	The trace simplifies to
	$\tr\{\mQ\mA\mSigma_q\mA\tran\}{\,=\,}\tr\{\bar{\mA}\tran\bar{\mA}\mSigma_q\}{\,=\,}d_z / \lambda$.
	Therefore, to satisfy the expectation constraint, $\lambda{=}d_z/\hat{h}$.
\end{proof}

\vspace{-1em}
\subsection{Additional Results}
\vspace{-1em}

\begin{figure}[!htb]
	{
		\input{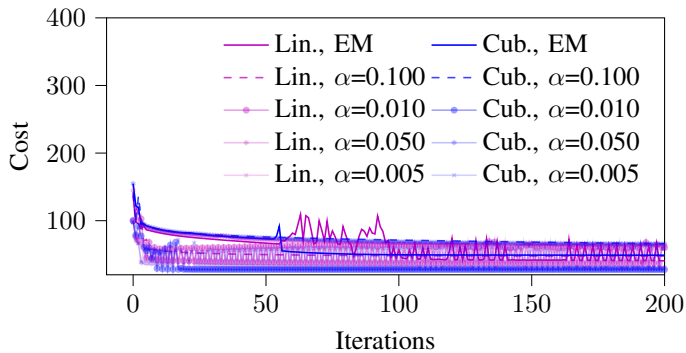}
		\vspace{-2em}
		\caption{Revisiting the experiment of Figure \ref{fig:dcp_cost} to compare approximate inference against $\alpha$ strategy. As illustrated in Section \ref{sec:optimization}, a fixed $\alpha$ can often lead to sub-optimal convergence or numerical stability.
		EM-based adaptive tuning automatically scales the hyperparameter to aid stability and performance.
		As illustrated in this task, careful optimization of a fixed $\alpha$ can result in superior performance when combining aggressive optimization with more accurate inference.}
		\label{fig:dcp_cost_alpha}
	}
\end{figure}

\newpage

\begin{IEEEbiography}[{\includegraphics[width=1in,height=1.25in,clip,keepaspectratio]{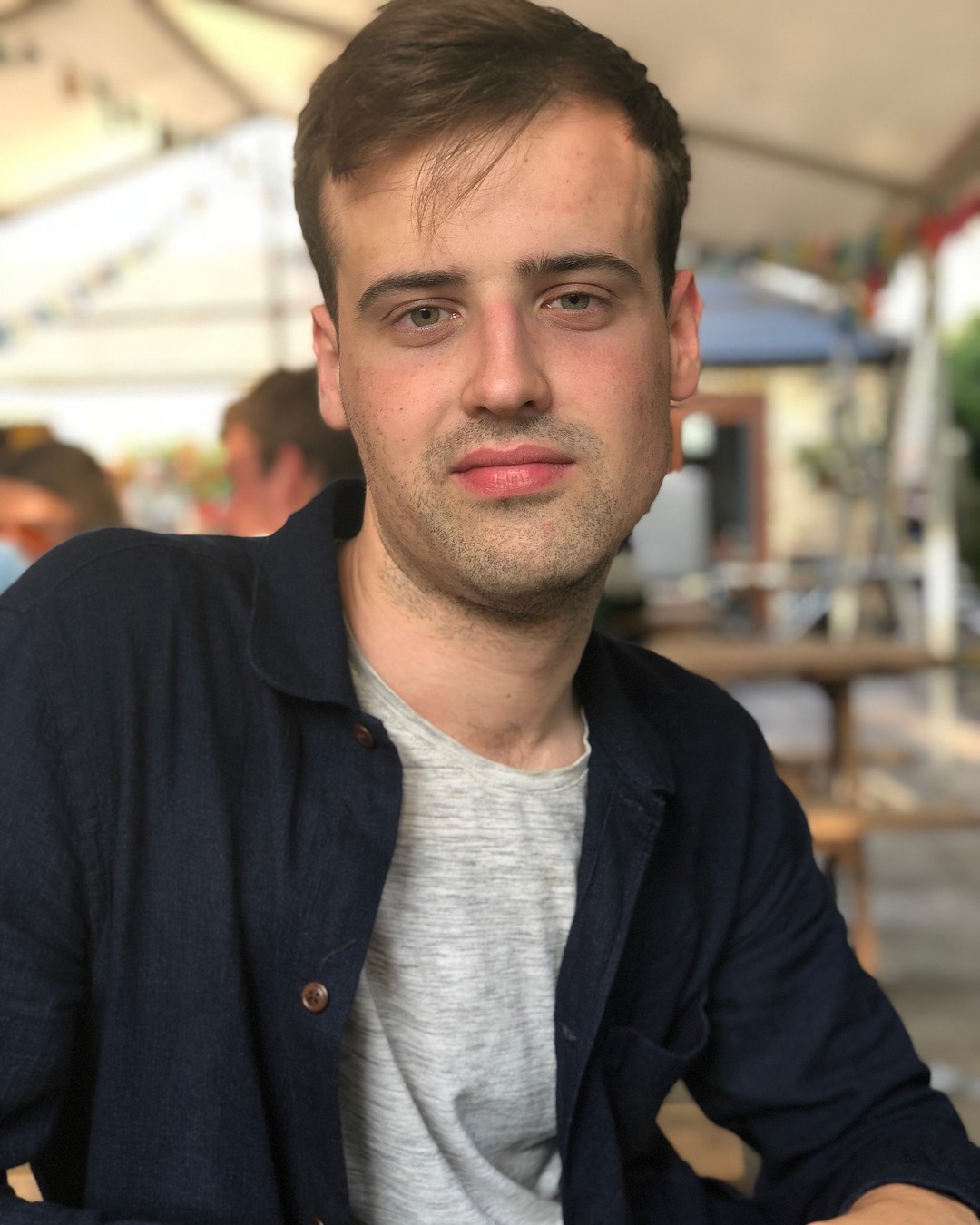}}]{Joe Watson}
received a BA and MEng in Information and Computer Engineering from Peterhouse, University of Cambridge in 2016, where he was the Charles Babbage Senior Scholar.
He is currently working towards a PhD degree with the Intelligent Autonomous Systems Group, Computer Science Department, Technical University of Darmstadt.
He is broadly interested in the intersection of control and probabilistic inference for sample-efficient learning algorithms for robotics.
\end{IEEEbiography}

\vspace{-1em}

\begin{IEEEbiography}[{\includegraphics[width=1in,height=1.25in,clip,keepaspectratio]{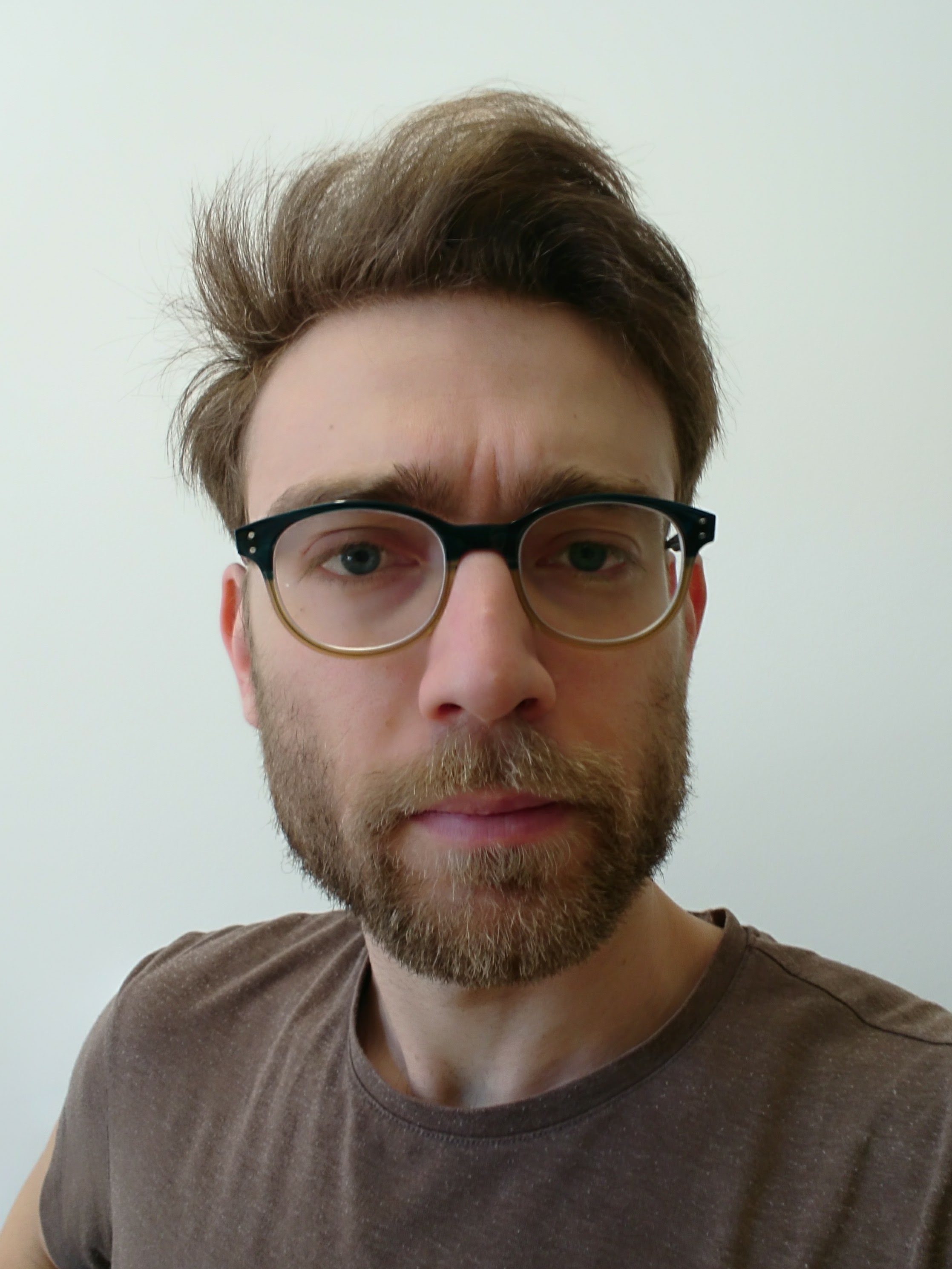}}]{Hany Abdulsamad}
is a postdoctoral researcher at Aalto University and the Finnish Center for Artificial Intelligence 
He completed his PhD degree with the Intelligent Autonomous Systems Group, Computer Science Department, Technical University of Darmstadt.
Hany's research considers the intersection of statistical inference and control, focusing on switching linear dynamics and hybrid systems.
\end{IEEEbiography}

\vspace{-2em}
\begin{IEEEbiography}[{\includegraphics[width=1in,height=1.25in,clip,keepaspectratio]{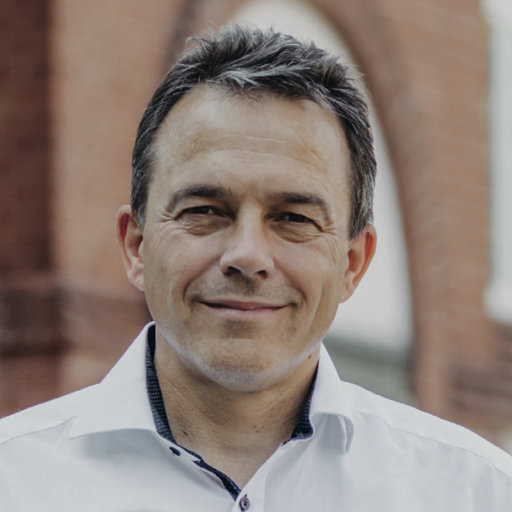}}]{Rolf Findeisen}
	is a full professor and heads the Control and CyberPhysical Systems Laboratory at the Technical University of Darmstadt.
	He received the M.S. degree from the University of Wisconsin, Madison, and the Ph.D. degree from the University of Stuttgart.
	Rolf was a research assistant in the Automatic Control Laboratory, ETH Z\"urich, and had several research stays and guest professorships, including Massachusetts Institute of Technology, Cambridge, USA, EPF Lausanne, Imperial College London.  
	Before moving the TU Darmstadt, he headed the Systems Theory and Control Laboratory at the Otto von Guericke University Magdeburg.
	His research interests focus on the model predictive control, the fusion of control and learning with guarantees, control of interconnected systems, and cyber-physical and network-controlled systems.
	The main fields of applications span mechatronics, robotics, autonomous driving to synthetic biology.
	Dr. Findeisen was the IPC Chair of the IFAC World Congress 2021.
	He has been an editor and associate editor for several journals, including the IEEE Transactions on Control of Network System and the IEEE Control Systems Magazine.  
\end{IEEEbiography}

\begin{IEEEbiography}[{\includegraphics[width=1in,height=1.25in,clip,keepaspectratio]{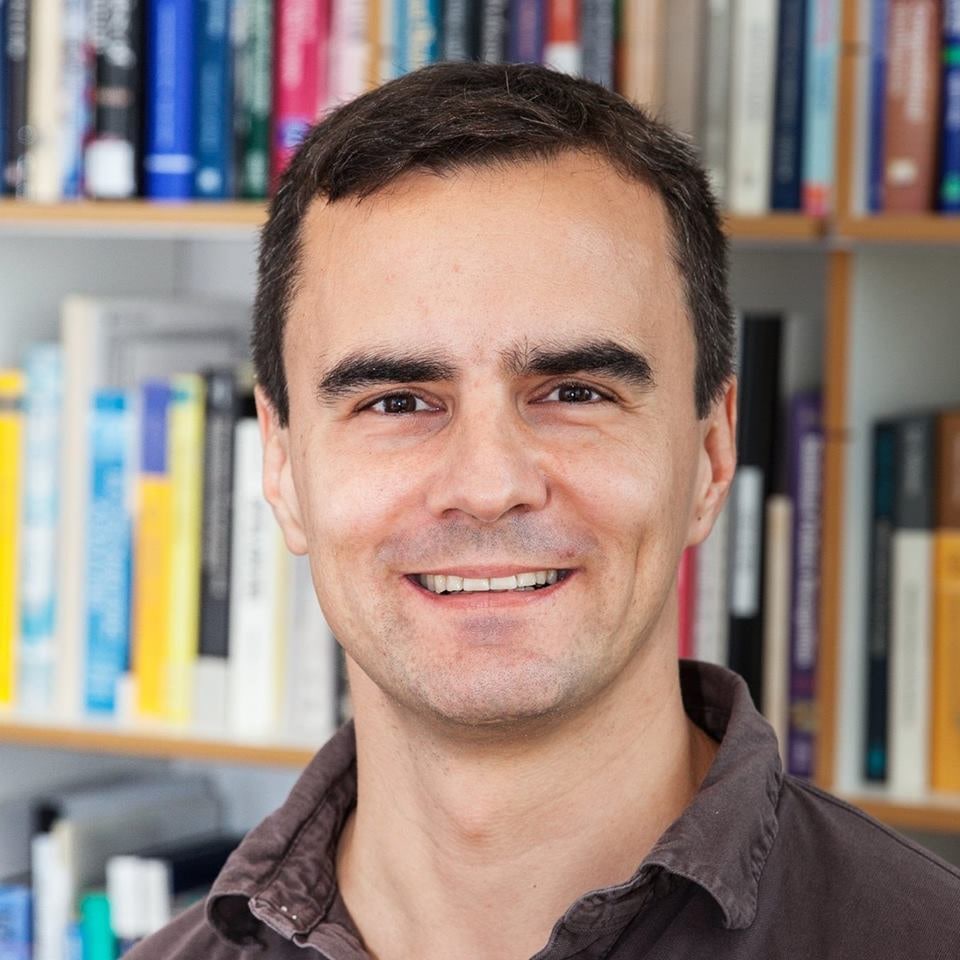}}]{Jan Peters}
	is a full professor (W3) for Intelligent Autonomous Systems at the Computer
	Science Department of the Technical University
	of Darmstadt.
	Jan Peters has received the Dick Volz
	Best 2007 US Ph.D. Thesis Runner-Up Award, the Robotics: Science \& Systems - Early Career Spotlight, the INNS Young Investigator Award, and the IEEE Robotics \& Automation Society’s Early Career Award as well as numerous best paper awards. In 2015, he received an ERC Starting Grant and in 2019, he was appointed as an IEEE Fellow.
\end{IEEEbiography}

\vfill\clearpage

\end{document}